\documentclass[11pt]{article}

\usepackage{microtype}
\usepackage{graphicx}
\usepackage{hyperref}

\usepackage{amsmath}  % For math
\usepackage{amssymb} 
\usepackage{amsthm}
\usepackage{subcaption}
\usepackage{url}
\usepackage{thm-restate}
\usepackage{dsfont}
\usepackage[group-separator={,}]{siunitx}
\usepackage[colorinlistoftodos]{todonotes}
\usepackage{fullpage}
\usepackage[square, numbers]{natbib}

\newcounter{example}

\newtheorem{definition}{Definition}
\newtheorem{assumption}{Assumption}

\newtheorem{corollary}{Corollary}

\newtheorem{goal}{Problem}
\newtheorem*{thm*}{Theorem}

\newcommand{\Xcal}{\ensuremath{\mathcal{X}}}
\newcommand{\Ycal}{\ensuremath{\mathcal{Y}}}

\newcommand{\EE}{\mathbb{E}}

\newcommand{\PP}{\mathbb{P}}
\newcommand{\QQ}{\mathbb{Q}}

\newcommand{\RR}{\mathbb{R}}

\newcommand{\indi}{\mathds{1}}

\newcommand{\p}[1]{\mu_{#1}}
\newcommand{\q}[1]{\tilde{\mu}_{#1}}
\newcommand{\xq}[1]{{q}_{#1}}
\newcommand{\tv}[2]{\operatorname{TV}(#1, #2)}

\newcommand{\nummodels}{h}
\newcommand{\nameset}{ordered}

\title{Why do classifier accuracies show linear trends\\ under distribution shift?}
\author{Horia Mania \qquad Suvrit Sra \\
{\normalsize Department of Electrical Engineering and Computer Science}\\
{\normalsize Massachusetts Institute of Technology}\\
{\normalsize \{hmania, suvrit\}@mit.edu}}
\date{December 30, 2020. Revised: February 22, 2021.}

\begin{document}
\maketitle

\begin{abstract}
Recent studies of generalization in deep learning have observed a puzzling trend: accuracies of models on one data distribution are approximately linear functions of the accuracies on another distribution. We explain this trend under an intuitive assumption on model similarity, which was verified empirically in prior work. 
More precisely, we assume the probability that two models agree in their predictions is higher than what we can infer from their accuracy levels alone. Then, we show that a linear trend must occur when evaluating models on two distributions unless the size of the distribution shift is large. 
This work emphasizes the value of understanding model similarity, which can have an impact on the generalization and robustness of classification models.   
\end{abstract}

\section{Introduction}
An important question that arises when evaluating deep learning models is whether the community has overfit to the test sets of popular datasets. Recent studies address this question by trying to replicate the original data collection pipelines and then evaluate models on the newly collected data  \cite{miller2020effect, recht2019imagenet,roelofs2019meta,yadav2019cold}.   

\begin{figure}[h]
\centering
\begin{subfigure}[b]{.45\linewidth}	
\includegraphics[width=\linewidth]{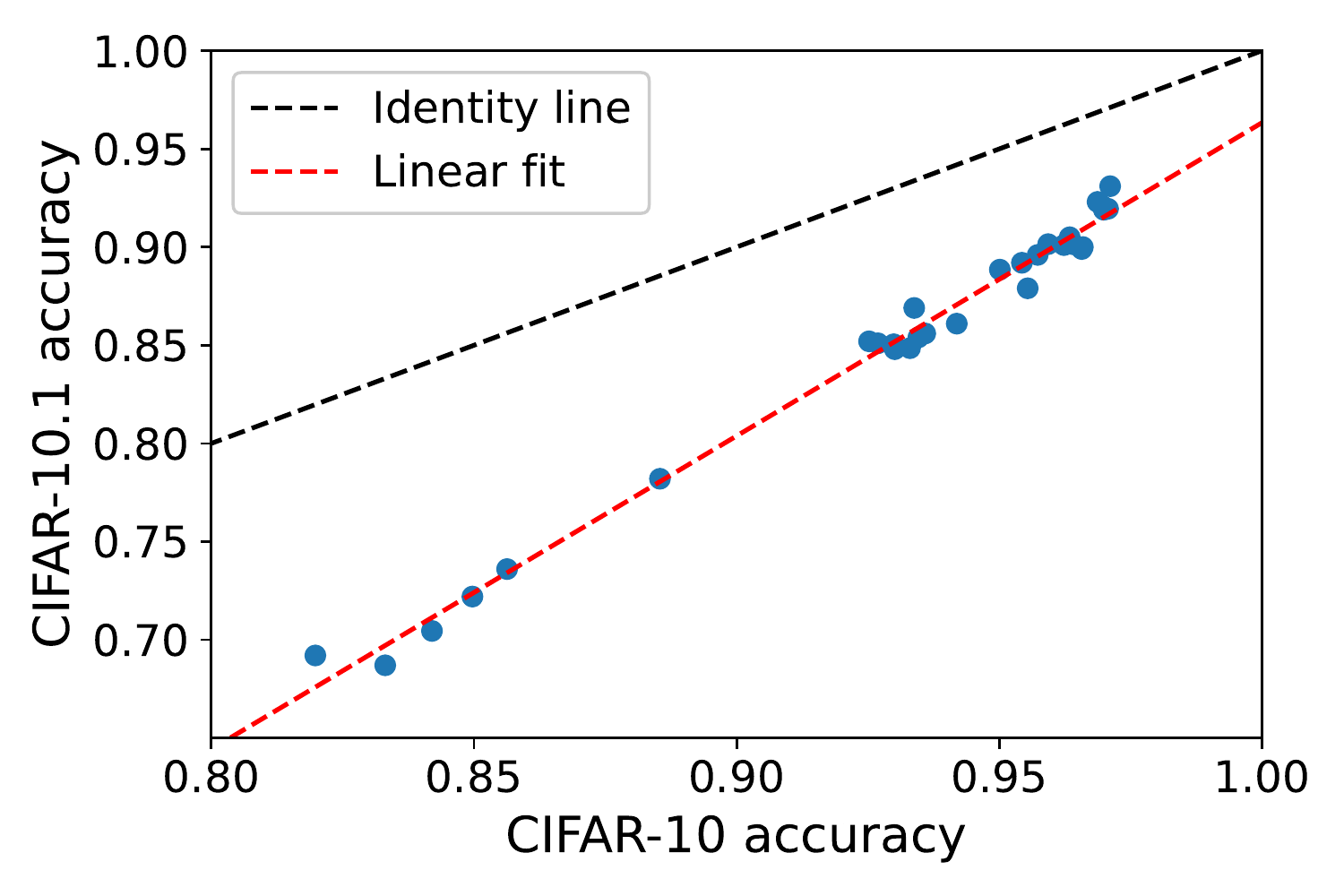}
\end{subfigure}
\centering
\begin{subfigure}[b]{.45\linewidth}
\includegraphics[width=\linewidth]{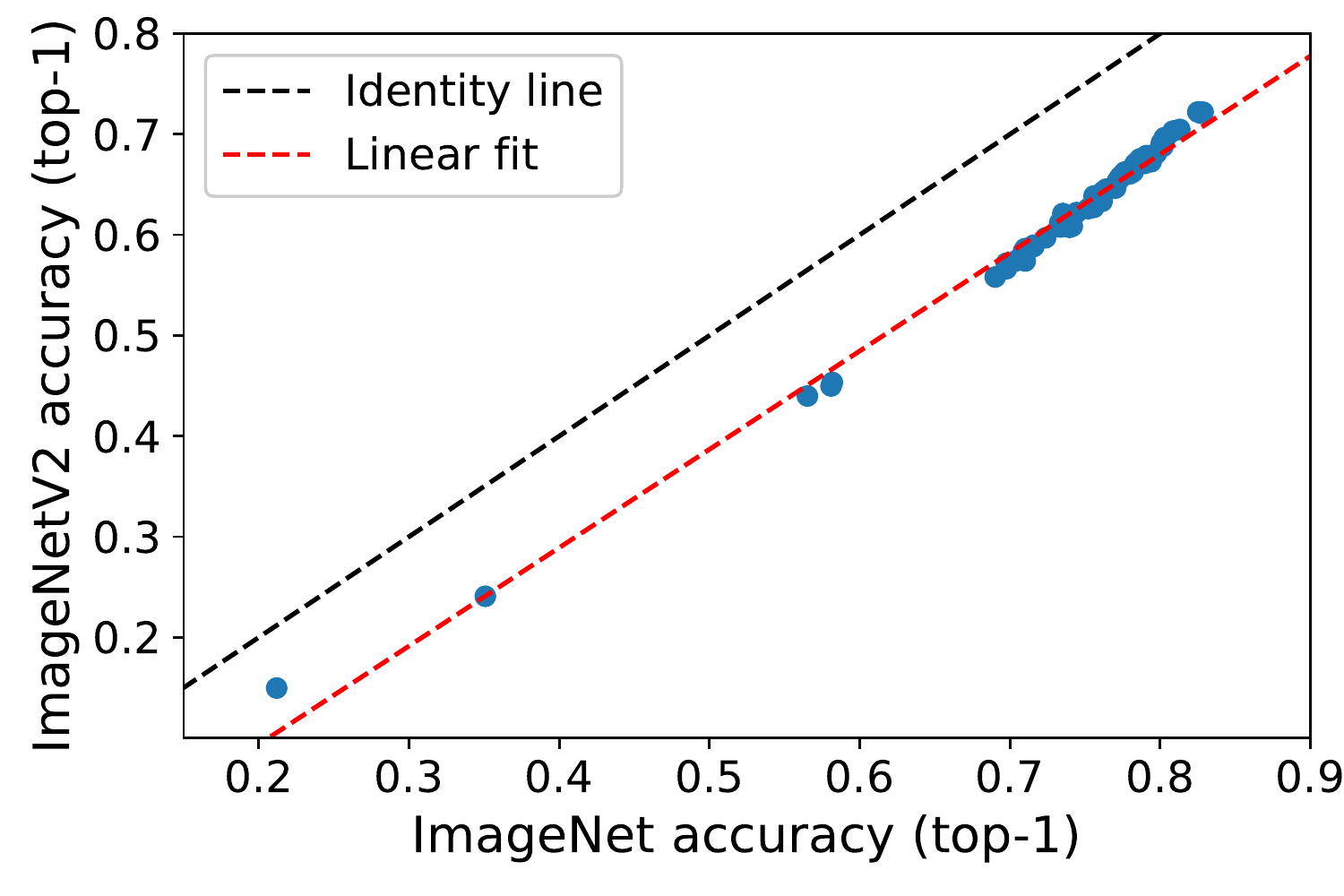}
\end{subfigure}	

\caption{These two figures summarize the findings of \citet{recht2019imagenet}. The blue points in the left plot represent $30$ CIFAR-10 models evaluated on both the original test set and a new test set, known as CIFAR-10.1. The right plot shows $66$ ImageNet models evaluated on both the original validation set and a new test set, known as ImageNetV2. For more examples of distributions shifts, with and without linear trends, one should visit \url{robustness.imagenetv2.org} to visualize the testbed developed by \citet{taori2020measuring}.
} 
\label{fig:example}
\end{figure}

While these studies find no overfitting, some of them inadvertently induce a shift between the distributions of the original and new test sets that leads to a drop in model performance. \citet{recht2019imagenet} observe accuracy drops of $3\% -\! 15\%$ on CIFAR-10 \cite{krizhevsky2009learning} and of $11\%\! -\! 14\%$ on ImageNet \cite{deng2009imagenet}. 
Their observations are summarized in Figure~\ref{fig:example}, from which 
one can notice a surprising pattern: Classification accuracies on new data are approximately linear functions of the accuracies on the original data, a phenomenon that has also been observed on question-answering data \cite{miller2020effect} and other variants of ImageNet \cite{barbu2019objectnet,taori2020measuring}. These observations raise the following key questions:
\begin{list}{\stepcounter{enumi}\textbf{\emph{Q\theenumi}.}}{\leftmargin=2.5em}
  \setlength{\itemsep}{0pt}
  \item \emph{Why are classification models approximately collinear when evaluated on two data distributions?}
  \item \emph{When can one expect this phenomenon to occur?}
\end{list}

Answers to these questions provide insights into improving model robustness
because training a model robust to distribution shift would mean training a
model that is \emph{not} collinear with models developed previously; it would mean training a model that lies close to the identity lines (black dashed lines) in Figure~\ref{fig:example}. Therefore, to build robust models it will be helpful to first understand why the models available today are approximately collinear.

While a growing body of work is trying to build robust models \cite{ben2013robust,delage2010distributionally,duchi2019distributionally,esfahani2018data, sagawa2019distributionally, shafieezadeh2015distributionally,sinha2017certifying}, as far as we know, there are no models that match human robustness on CIFAR-10.1 and ImageNetV2, the datasets collected by \citet{recht2019imagenet} \cite{shankar2020evaluating,taori2020measuring}. 
We hope that understanding when and why models are approximately collinear would inform future work on robust machine learning. 

Driven by the above motivation we seek to answer questions \textbf{Q1} and \textbf{Q2}. We start from a premise inspired by the work of~\citet{mania2019model} on model similarity. They observe that the probability that two classifiers correctly classify or incorrectly classify a data point is larger than what one would expect if the two classifiers were making predictions independently. We translate this observation into the following assumption: \emph{given two models, the probability that the lower accuracy model classifies a data point correctly while the higher accuracy model classifies it incorrectly is small}. In other words, we assume that high accuracy models correctly classify most of the data points correctly classified by lower accuracy models. Notably, this assumptions is satisfied by ImageNet and CIFAR-10 models.

Under this assumption we show that models must be approximately collinear when evaluated on two distributions, unless the size of the distribution shift is large in a certain sense. 
We also discuss refinements of our analysis that explain why a probit axis scaling leads to an even better linear fit for ImageNet models, as observed by~\citet{recht2019imagenet}.

\section{Main results}
\label{sec:statement}
Let $\Xcal$ be a covariate space and $\Ycal$ a discrete label space. Let $\PP$ and $\QQ$ be two probability distributions on $\Xcal \times \Ycal$. Also, suppose we have a set of $\nummodels$ models $f_1$, $f_2$, \ldots, $f_\nummodels$ that map $\Xcal$ to $\Ycal$. We are interested in the accuracies
\begin{align*}
  \p{i} &:= \EE_\PP \indi(f_i(x) = y),\quad\text{and} \quad \q{i} := \EE_\QQ \indi(f_i(x) = y),
\end{align*}
where the expectations are with respect to the data point $(x,y) \in \Xcal \times \Ycal$, distributed according to $\PP$ and $\QQ$, respectively. Given this notation, our main goal is to show that the accuracies $\{(\p{i}, \q{i})\}_{i = 1}^\nummodels$ are approximately collinear, which can be formulated more precisely as follows:
\begin{goal}
Under suitable assumptions, show that there exist $\alpha, \beta \in \RR$ such that $\q{i}\approx \alpha \p{i} + \beta$ for $1\leqslant i\leqslant \nummodels$. Alternatively, given a distribution $\PP$ and a set of models $f_1$, $f_2$, \ldots, $f_\nummodels$ , show that for a large class of distribution shifts there exist $\alpha, \beta \in \RR$ such that $\q{i}\approx \alpha \p{i} + \beta$ for $1\leqslant i\leqslant \nummodels$.
\label{prob:main_goal}
\end{goal}

Our main assumption for Problem~\ref{prob:main_goal} is drawn from~\citep{mania2019model}, who observe that the predictions of different image classification models are more similar than one would expect based on accuracies alone. Specifically, if two models $f_1$ and $f_2$ with accuracies $\mu_1$ and $\mu_2$ were making mistakes independently of each other, the probability that both models correctly or incorrectly classify a data point would be $\mu_1 \mu_2 + (1 - \mu_1) (1 - \mu_2)$. However, they empirically evaluate these similarity probabilities for ImageNet and CIFAR-10 models and observe that model similarities are significantly larger than what the similarities would be if the models were making mistakes independently.

Model similarity is difficult to work with for our purposes because a similarity value can be said to be high or low only in relation to the models' accuracies, which in our analysis vary between $21\%$ and $83\%$. Instead we work with a more suitable quantity related to model similarity. 

\begin{assumption}
\label{as:dom}
For any pair of models $f_i$ and $f_j$ with $\p{i} \leq \p{j}$, we have
\begin{align*}
\PP(\{f_i(x) = y\} \cap \{f_j(x) \neq y\}) \leq \zeta.
\end{align*}
\end{assumption}
We postpone a discussion of this assumption to Section~\ref{sec:similarity}, where we show empirically that most probabilities $\PP(\{f_i(x) = y\} \cap \{f_j(x) \neq y\})$ for CIFAR-10 and ImageNet models are less than $0.05$. When a set of models satisfies Assumption~\ref{as:dom} with $\zeta = 0$ we call it \emph{\nameset}, and if a pair of models $f_i$, $f_j$ satisfies Assumption~\ref{as:dom} with $\zeta=0$, we say that 
$f_j$ \emph{dominates} $f_i$. More generally, we call $\PP(\{f_i(x) = y\} \cap \{f_j(x) \neq y\})$ the \emph{dominance probability}. 

Regardless of $\zeta$, Assumption~\ref{as:dom} by itself is insufficient to guarantee that classification accuracies lie close to a line. In fact, in Section~\ref{sec:examples} we show that even when $\zeta = 0$ and Assumption~\ref{as:dom} holds for both $\PP$ and $\QQ$, there still exist models that are \emph{not} approximately collinear. To ensure collinearity, we need another ingredient, namely, suitable closeness of the distributions $\PP$ and $\QQ$; we present our notion below.

\subsection{Characterizing distributional closeness}

Various notions of distance between distributions exist, each suitable for different applications~\citep{deza}. 
One of the most well-known distances is the total variation: $\tv{\PP}{\QQ} = \sup_A |\PP(A) - \QQ(A)|$, where the supremum is taken with respect to all measurable events. This distance is stringent because it requires $|\PP(A) - \QQ(A)|$ to be small for all $A$. However, for our analysis it suffices to measure $|\PP(A) - \QQ(A)|$ for only $6 \binom{\nummodels}{3}$ events that depend on the set $\{f_1, f_2, \ldots, f_\nummodels\}$ of models.

To see why $6 \binom{\nummodels}{3}$ events suffice, we first need some notation. For  a model $f_i$ let $A_i^+$ denote the subset of $\Xcal \times \Ycal$ on which $f_i$ is correct, and $A_i^-$ the subset on which $f_i$ is incorrect. Now we are ready to introduce our notion of closeness. %between $\PP$ and $\QQ$ that we require.  

\begin{definition}
  We say $\PP$ and $\QQ$ are $(\delta_1, \delta_2, \nu_1, \nu_2)$-close if 
  for all distinct $i,j,k \in \{1,2,\ldots, \nummodels\}$ and $(\varepsilon_i, \varepsilon_j, \varepsilon_k) \in \{-,+\}^3$, different from $(-,-,-)$ and $(+,+,+)$, we have
  \begin{equation}
    \begin{split}
      - \nu_1 + & (1 - \delta_1) \PP(A_i^{\varepsilon_i} \cap A_j^{\varepsilon_j} \cap A_k^{\varepsilon_k}) \leq \QQ(A_i^{\varepsilon_i} \cap A_j^{{}\varepsilon_j} \cap A_k^{\varepsilon_k}) \leq \nu_2 + (1 + \delta_2) \PP(A_i^{\varepsilon_i} \cap A_j^{\varepsilon_j} \cap A_k^{\varepsilon_k}).
      \label{eq:as_refined}
    \end{split}
  \end{equation}
  \label{def:dist}
\end{definition}

In Section~\ref{sec:size_dist_shift} we provide empirical motivation for this definition and a more detailed discussion. For now we make a few remarks. Since \eqref{eq:as_refined} imposes constraints on the probabilities $\QQ(A)$ only for $6 \binom{\nummodels}{3}$ events $A$, the distance between $\QQ$ and $\PP$ is allowed to be large when measured using traditional distances such as $\operatorname{TV}$ or $\operatorname{KL}$. In fact, the smaller the number of events $A$ for which $|\PP(A) - \QQ(A)|$ has to be small the easier it is for $\PP$ and $\QQ$ to be close. 

Also, $\PP$ and $\QQ$ being $(\delta_1, \delta_2, \nu_1, \nu_2)$-close 
does not guarantee by itself that the models $f_1$, $f_2$, \ldots, $f_\nummodels$ are approximately collinear, unless $\delta_1$, $\delta_2$, $\nu_1$ and $\nu_2$ are much smaller than the values we observe in practice. 
In Section~\ref{sec:size_dist_shift} we show that the empirical distributions of ImageNet and ImageNetV2 are $(0.31, 0.38, 0.005, 0.008)$-close, and in Section~\ref{sec:examples} we show that it is possible to have three classification models and two distributions that are $(0.31, 0.38, 0.005, 0.008)$-close but the three models are far from collinear.

While neither Assumption~\ref{as:dom} nor $(\delta_1, \delta_2, \nu_1, \nu_2)$-closeness is sufficient to guarantee that models are approximately collinear, we show that together they are sufficient. The main idea of our solution to Problem~\ref{prob:main_goal} is to look at any three models $f_i$, $f_j$, and $f_k$ with accuracies $\p{i} \leq \p{j} \leq \p{k}$, and to consider the line $\ell$ determined by the points $(\p{i}, \q{i})$ and $(\p{k}, \q{k})$. Then, we upper bound the residual  
from $(\p{j}, \q{j})$ to the line $\ell$, i.e., we upper bound $|\ell(\p{j}) - \q{j}|$. The next result encapsulates our analysis. 

\begin{restatable}{prop}{main}
Let $f_1$, $f_2$, \ldots, $f_\nummodels$ be an \nameset{} set of models and $\PP$,  $\QQ$ two distributions that are $(\delta_1, \delta_2, \nu_1, \nu_2)$ close. Then, for any three models $f_i$, $f_j$, $f_k$ with $\p{i} \leq \p{j} \leq \p{k}$, if $\ell$ is the line defined by the points $(\p{i}, \q{i})$ and $(\p{k}, \q{k})$ and $r$ is the residual $|\ell(\p{j}) - \q{j}|$, we have 
\begin{equation}
\label{eq:basic_bound}
\begin{split}
r &\leq 
\frac{\delta_1 + \delta_2}{2} \frac{2(\p{k} - \p{j})(\p{j} - \p{i})}{\p{k} - \p{i}} + \max\{\nu_1, \nu_2\} + \left(1 + \frac{\max\{\p{k} - \p{j}, \p{j} - \p{i}\}}{\p{k} - \p{i}}\!\right) \nu_2.
\end{split}
\end{equation}
Moreover, for any three models there exists a line $\ell^\prime$ such that all three residuals are equal to $r^\prime = r/2$. 
\label{prop:main}
\end{restatable}
The main takeaway of this result is that given an \nameset{} set of models $f_1$, $f_2$, \ldots, $f_\nummodels$ with respect to a distribution $\PP$, the models are guaranteed to be approximately collinear when evaluated on a distribution $\QQ$, unless the distribution shift between $\PP$ and $\QQ$ is large in the sense of \eqref{eq:as_refined}. 

\begin{figure}[t]
\centering
\begin{subfigure}[b]{.45\linewidth}
\includegraphics[width=\linewidth]{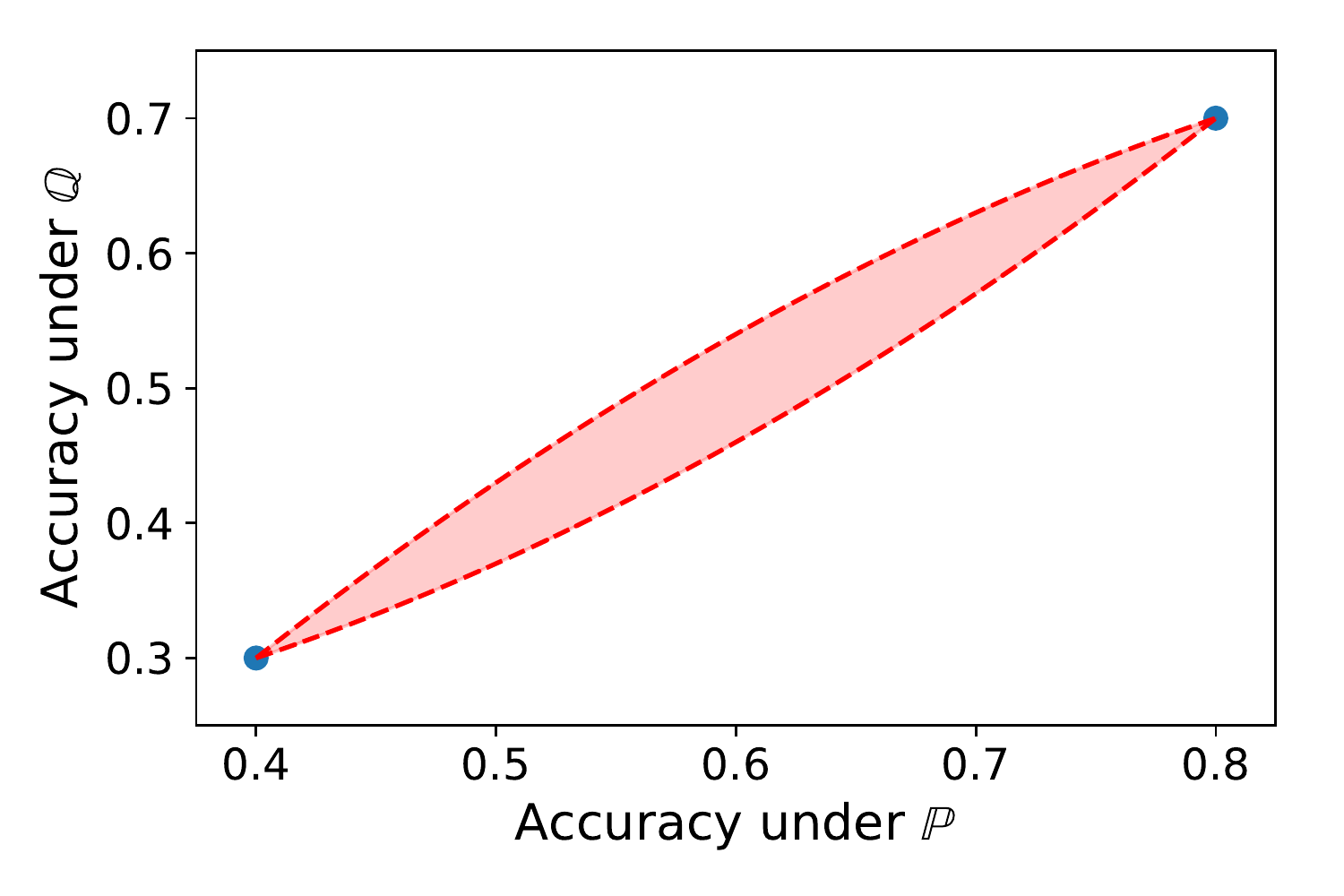}
\caption{Guarantee after applying \eqref{eq:basic_bound} once.}
\label{fig:hypotheticalv1}
\end{subfigure}
\begin{subfigure}[b]{.45\linewidth}
\includegraphics[width=\linewidth]{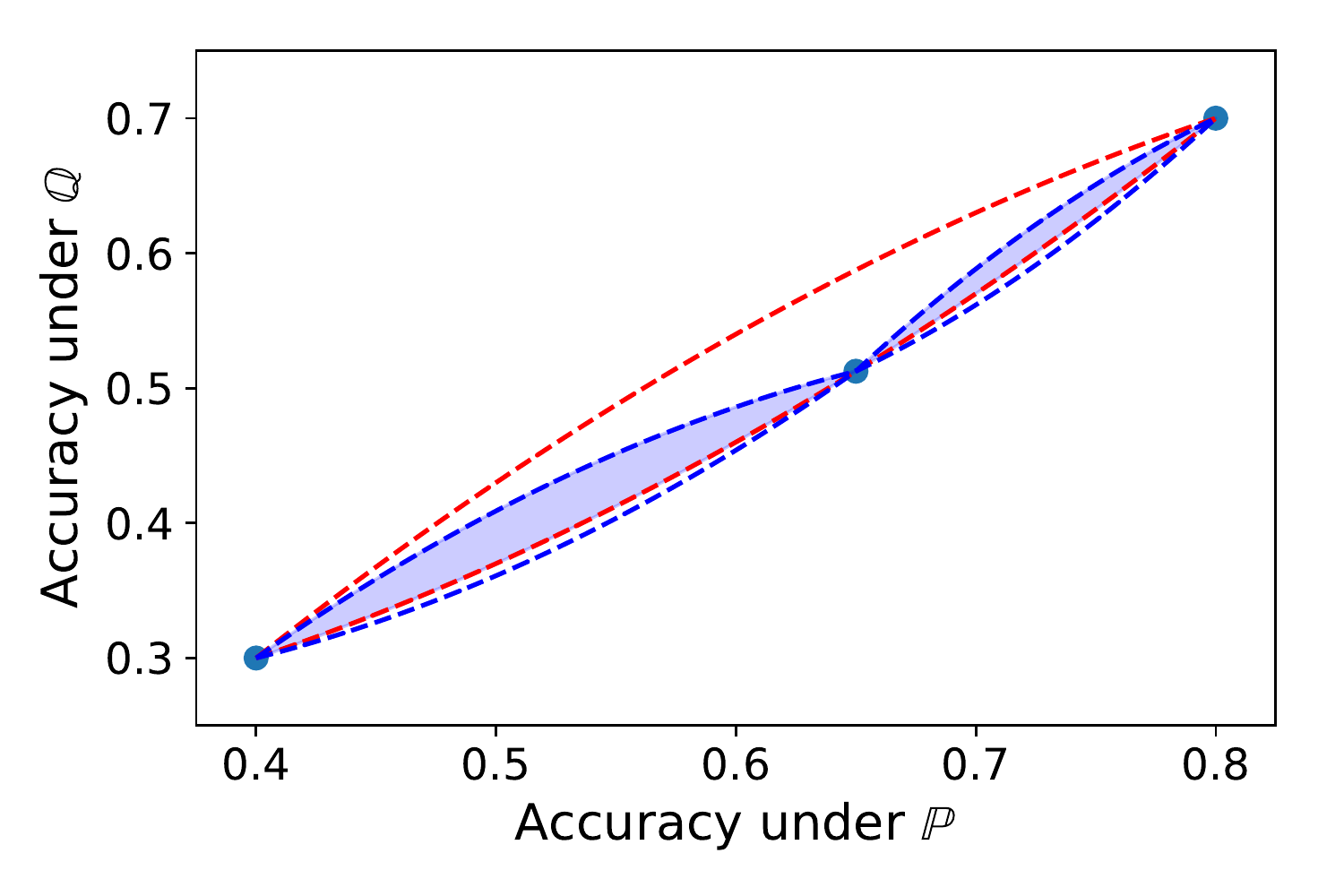}
\caption{Guarantee after applying \eqref{eq:basic_bound} thrice.}
\label{fig:hypotheticalv2}
\end{subfigure}
\caption{A depiction of bound \eqref{eq:basic_bound}. The blue points represent hypothetical models. Any other models must lie inside the shaded regions when the following conditions hold: the two distributions are $(0.2, 0.2, 0.0, 0.0)$-close and the set of models is \nameset{}. 
}
\label{fig:viz_bounds}
\end{figure}

Notice that the second factor in the first term of \eqref{eq:basic_bound} is the harmonic mean of $\p{k} - \p{j}$ and $\p{j} - \p{i}$. To gain intuition about this bound, Figure~\ref{fig:hypotheticalv1} depicts its guarantee for three ordered models when $\delta_1 = \delta_2 = 0.2$ and $\nu_1 = \nu_2 = 0$. More concretely, we consider two hypothetical models $f_i$ and $f_k$ with $(\p{i}, \q{i}) = (0.4, 0.3)$ and $(\p{k}, \q{k}) = (0.8, 0.7)$. Then, Proposition~\ref{prop:main} ensures that any other model $f_j$ that dominates $f_i$ and is dominated by $f_k$ must lie in the red shaded region of Figure~\ref{fig:hypotheticalv1}. Therefore, if we were to draw the line defined by the two blue points, all other models in the ordered set would lie close to it. 

Moreover,  \eqref{eq:basic_bound} applies to any subset of three models. Figure~\ref{fig:hypotheticalv2} shows what kind of guarantee we can obtain when we take advantage of this fact. Suppose $f_j$ is the middle blue point shown in Figure~\ref{fig:hypotheticalv2}. Then, by applying \eqref{eq:basic_bound} two more times we see that a fourth model would have to lie in the smaller blue shaded region. Figure~\ref{fig:hypotheticalv2} shows that in cases when a linear fit produces residuals that are larger than desired, one can expect to see a much better fit when using piecewise linear regression with two pieces. We come back to this point in Section~\ref{sec:probit}, when we discuss why using probit scaling for plotting the ImageNetV2 accuracies as a function of ImageNet accuracies improves the linear fit. 

Proposition~\ref{prop:main} is the core of our answer to Problem~\ref{prob:main_goal}. It shows that any three models must be approximately collinear when the set of models is \nameset{} and the distribution shift is not too large, as measured by \eqref{eq:as_refined}. The next corollary shows that all models must be approximately collinear; the proof is deferred to Appendix~\ref{app:cor}.

\begin{corollary}
Let $f_1$, $f_2$, \ldots, $f_\nummodels$ be an \nameset{} set of models and $\PP$,  $\QQ$ be two distributions that are $(\delta_1, \delta_2, \nu_1, \nu_2)$-close.
Then, there exists a line such that the residual from any point $(\p{i}, \q{i})$ to it is at most
\begin{equation}
\frac{25}{64}(\p{\nummodels} - \p{1}) \frac{\delta_1 + \delta_2}{2} + 3\max\{\nu_1, \nu_2\}.
\end{equation}
\label{cor:main}
\end{corollary}
As expected, this result says that the smaller the distribution shift is (i.e., small $\delta_1$, $\delta_2$, $\nu_1$, and $\nu_2$), the closer to a line the models will be. Moreover, the upper bound on the residual depends on the maximum accuracy difference between models in the set. Therefore, if we partition the models into two or more sets, we can get a tighter guarantees on the residuals of a piecewise linear fit.

\subsection{Examples and Notation}
\label{sec:examples}

In this section we show that neither Assumption~\ref{as:dom} nor the $(\delta_1, \delta_2, \nu_1, \nu_2)$-closeness of $\PP$ and $\QQ$ is alone sufficient to guarantee that models are approximately collinear. 

We introduce some notation that is also used in subsequent sections. Given $f_1$, $f_2$, and $f_3$ we denote the probabilities
\begin{align*}
p_{123} &= \PP(A_1^+ \cap A_2^+ \cap A_3^+),\\
p_{23} &= \PP(A_1^- \cap A_2^+ \cap A_3^+),\\
p_{3} &= \PP(A_1^- \cap A_2^- \cap A_3^+).
\end{align*} 
In words, $p_{123}$ is the probability that all three models $f_1$, $f_2$, and $f_3$ classify a data point sampled from $\PP$ correctly. The term $p_{23}$ is the probability that models $f_2$ and $f_3$ classify a  data point correctly while $f_1$ classifies it incorrectly. %By now it should be clear what $p_3$ represents. 
We also need to consider the probabilities: 
\begin{align*}
p_{1} &= \PP(A_1^+ \cap A_2^- \cap A_3^-), \quad p_{12} = \PP(A_1^+ \cap A_2^+ \cap A_3^-),\\
p_{2} &= \PP(A_1^- \cap A_2^+ \cap A_3^-), \quad p_{13} = \PP(A_1^+ \cap A_2^- \cap A_3^+).
\end{align*}
These four probabilities are zero when $f_1$, $f_2$, and $f_3$ are \nameset{}. Therefore, 
under the ordering assumption, we have 
\begin{align*}
\p{1} = p_{123}, \quad \p{2} = p_{123} + p_{23}, \quad \p{3} = p_{123} + p_{23} + p_3.
\end{align*}
We use $\xq{123}$, $\xq{12}$, $\xq{1}$ and so on to denote the analogous probabilities under $\QQ$. Note that although $p_1$ is zero, $\xq{1}$ can be nonzero. When the 
set of models is ordered and the distributions $\PP$ and $\QQ$ are $(\delta_1, \delta_2, \nu_1, \nu_2)$-close  we know that $\xq{1}$, $\xq{2}$, $\xq{12}$, and $\xq{13}$ are at most $\nu_2$.

\paragraph{Example 1.} We show that there exist models $f_1$, $f_2$, and $f_3$ that are \nameset{} with respect to both $\PP$ and $\QQ$, but that are not approximately collinear. We emphasize that in this example the models are \nameset{} with respect to both $\PP$ and $\QQ$, which is a stronger requirement than that of Assumption~\ref{as:dom}. Regardless, this requirement is still insufficient to guarantee approximate collinearity of the models. 

Since the events $A_1^{\epsilon_1} \cap A_2^{\epsilon_2} \cap A_3^{\epsilon_3}$ are disjoint, there exist models $f_1$, $f_2$, $f_3$ and distributions $\PP$ and $\QQ$ such that these events have any probabilities we wish. We choose
\begin{align*}
p_{123} &= 0.6, \quad p_{23} =  0.1, \quad p_{3} = 0.1\\
p_{1} &= p_2 = p_{12} = p_{13} = 0. 
\end{align*}
With this choice the models $f_1$, $f_2$, $f_3$ have accuracies $0.6$, $0.7$, and $0.8$ and three models are \nameset{}. 

Now, we can choose the values of the probabilities of the same events under $\QQ$.
Since we want the models to be \nameset{} with respect to $\QQ$ too, we set $\xq{1} = \xq2 = \xq{12} = \xq{13} = 0$. However, without any constraint on the shift between $\PP$ and $\QQ$ we can set $\xq{123}$, $\xq{23}$, and $\xq{3}$ to be any nonnegative values whose sum is smaller or equal than one. For example, we can choose $\xq{123} = 0.5$, $\xq{23} = 0.4$, and $\xq{3} = 0$, which implies that $f_1$, $f_2$, and $f_3$ have accuracies $0.5$, $0.9$ and $0.9$ under $\QQ$. Therefore, the residual of $f_2$ from the line defined by $f_1$ and $f_2$ is $0.2$. This residual could be made even larger with a different choice of $\PP$ and $\QQ$. The takeaway from this example is that Assumption~\ref{as:dom} by itself does not guarantee approximate collinearity even when it holds with respect to both $\PP$ and $\QQ$. To avoid such a situation $\PP$ and $\QQ$ must be sufficiently close in some sense. 

\paragraph{Example 2.} The perceptive reader will notice that when $\PP$ and $\QQ$ are $(0,0,0,0)$-close the models must be  collinear. However, in Section~\ref{sec:size_dist_shift} we show that ImageNet and ImageNetV2 are $(0.31,0.38,0.005,0.008)$-close. In this example we show that for such a distribution shift there exist $f_1$, $f_2$, and $f_3$ that are far from being collinear. The main message of our analysis is that this situation \emph{cannot} occur when the models are \nameset{}. 

We consider models $f_1$, $f_2$, and $f_3$ with accuracies $\p{1}= 0.6$, $\p{2} = 0.7$, and $\p{3}= 0.8$. Then, we choose the probabilities of the events $A_1^{\epsilon_1} \cap A_2^{\epsilon_2} \cap A_3^{\epsilon_3}$ as if the models were making predictions independently:
\begin{align*}
p_{123} &= \p{1} \p{2} \p{3} = 0.336,\\
p_{1} &= \p{1} (1 - \p{2}) (1 - \p{3}) = 0.036,\\
p_{23} & = (1 - \p{1}) \p{2} \p{3} = 0.224,
\end{align*}
and $p_{12} = 0.084$, $p_{13} = 0.144$, $p_2 = 0.056$, and $p_{3} = 0.096$. These models are not ordered because the dominance probabilities are $p_1 + p_{13} = 0.18$, $p_1 + p_{12} = 0.12$, and $p_2 + p_{12} = 0.14$. Therefore, Assumption~\ref{as:dom} is satisfied only if $\zeta \geq 0.18$, which is twice as large as the value needed for ImageNet models (see Section~\ref{sec:dominance}). 

Now, if we choose $\xq{123} = 0.336$, $\xq{12} = 0.053$, $\xq{13} = 0.2$, $\xq{23} = 0.15$, $\xq{1} = 0.057$, $\xq{2} = 0.034$, and $\xq{3} = 0.14$, we get a distribution $\QQ$ that is $(0.31, 0.38, 0.005, 0.008)$-close to $\PP$. Under $\QQ$ the models $f_1$, $f_2$, and $f_3$ have accuracies $0.646$, $0.573$, and $0.826$, which means  that the residual of  
$f_2$ is $0.163$. On the other hand, if the models were \nameset{}, Proposition~\ref{prop:main} would guarantee that the residual is at most $0.055$. Therefore, in addition to $\PP$ and $\QQ$ being close, Assumption~\ref{as:dom} on model dominance needs to hold with a small $\zeta$ to guarantee that models are approximately collinear.

\section{Model dominance}
\label{sec:similarity}
In this section we take a closer look at Assumption~\ref{as:dom}. \citet{mania2019model} observe that image classification models make similar predictions and use this observation to show that one can re-use test sets more times than previously expected without overfitting. They define the similarity between two models $f_i$ and $f_j$ to be $\PP\left(\indi\{f_i(x) = y\} = \indi(f_j(x) = y)\right)$.

If two models were making mistakes independently of each other, this similarity would be $\p{i}\p{j} + (1 - \p{i})(1 - \p{j})$. However, on ImageNet they observed % \citet{mania2019model} observe
model similarities that are approximately $0.25$ higher.

\begin{figure}[t]
\centering
\begin{subfigure}[b]{.45\linewidth}
\includegraphics[width=\linewidth]{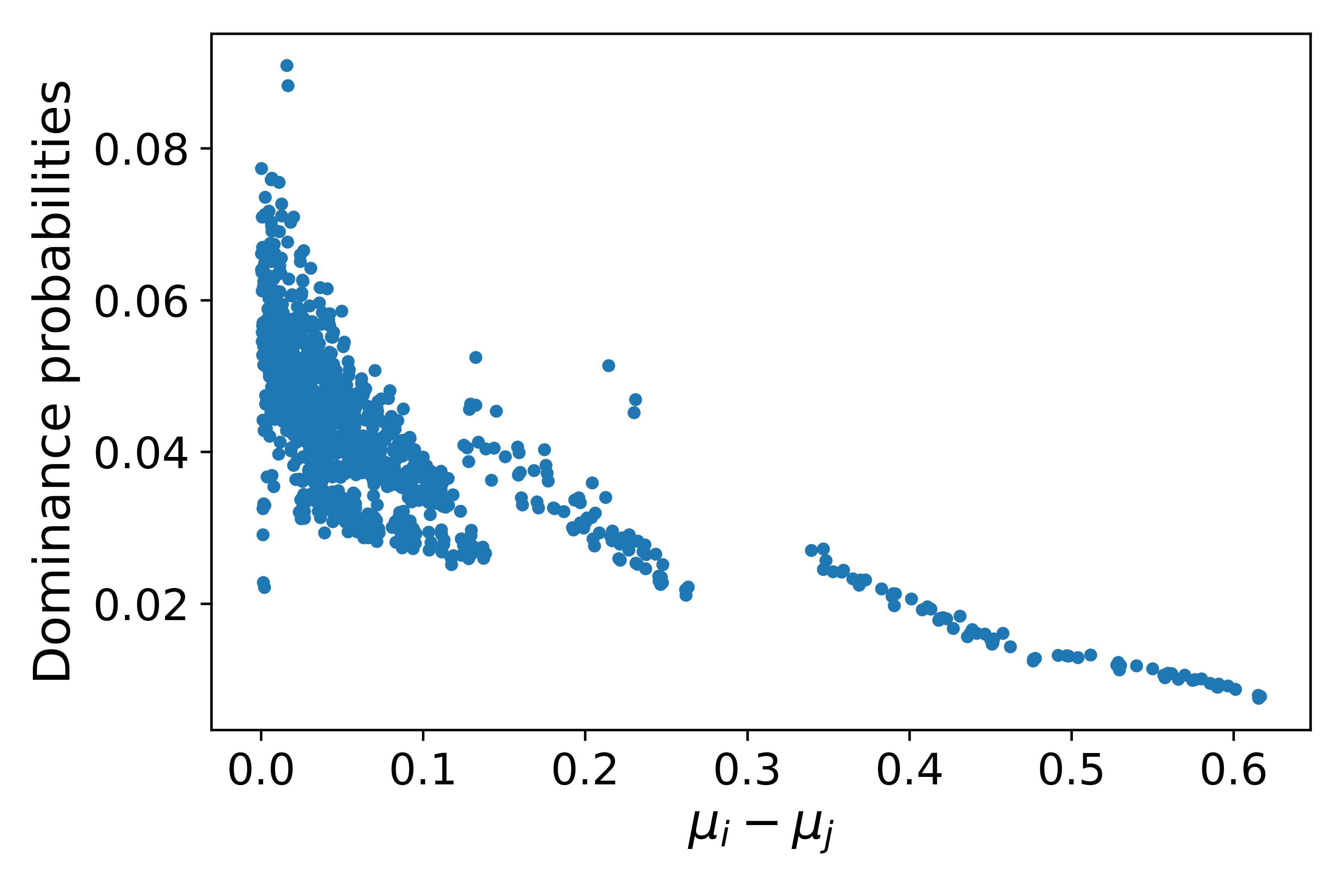}
\caption{Dominance probabilities}
\label{fig:dominance}
\end{subfigure}
\begin{subfigure}[b]{.45\linewidth}
\includegraphics[width=\linewidth]{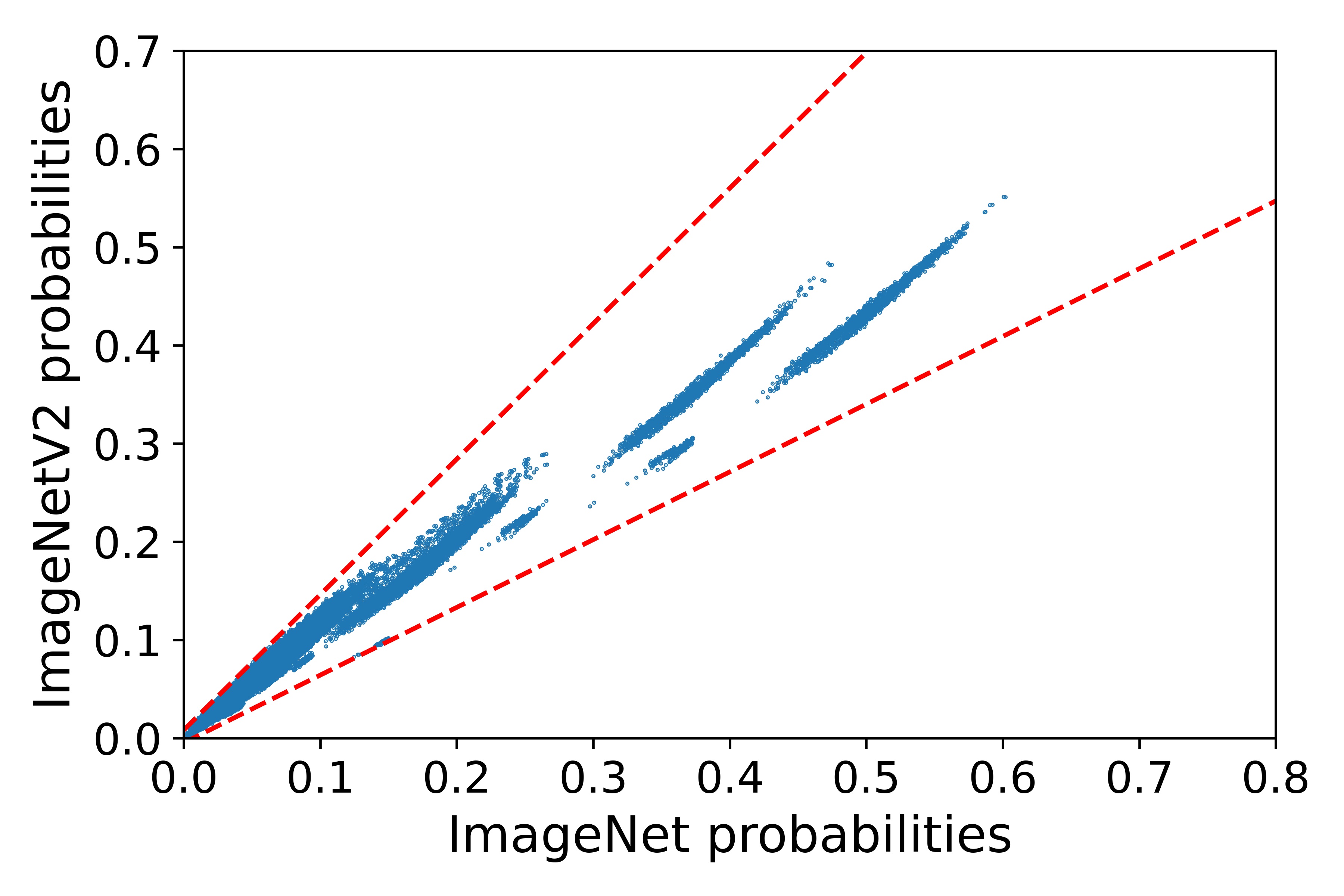}
\caption{Events defined by triplets}
\label{fig:triplets}
\end{subfigure}
\caption{Figure~\ref{fig:dominance} shows the dominance probabilities for all pairs of models. Figure~\ref{fig:triplets} shows $\num{274560}$ blue points representing the probabilities of the events $A_i^{\varepsilon_i} \cap A_j^{\varepsilon_j} \cap A_k^{\varepsilon_k}$. 
The red dotted lines are upper and lower bounds, as in \eqref{eq:as_refined}, with parameters: $\delta_1 = 0.31$, $\delta_2 = 0.38$, $\nu_1 = 0.005$, and $\nu_2 = 0.008$.}
\label{fig:as_evidence}
\end{figure}

We explained in Section~\ref{sec:statement} that it is a bit difficult to work with model similarity for our purposes. Instead, given two models $f_i$ and $f_j$ with $\p{i} \leq \p{j}$
we consider the probability that $f_i$ classifies  a data point correctly while $f_j$ does not:
\begin{align}
\label{eq:dominance_prob}
\PP(\{f_i(x) = y\} \cap \{f_j(x) \neq y\}).
\end{align}
Recall that we refer to \eqref{eq:dominance_prob} as dominance probability. It is easy to relate the model similarity and dominance probabilities since the former is equal to
\begin{align*}
1 - (\p{j} - \p{i}) - 2 \PP(\{f_i(x) = y\} \cap \{f_j(x) \neq y\}). 
\end{align*}
Assumption~\ref{as:dom} can be checked empirically. We study $\nummodels = 66$ ImageNet models that were collated and evaluated by \citet{recht2019imagenet} \footnote{\citet{recht2019imagenet} provide $67$ models. Even though all $67$ models are approximately collinear, we take out one of the three Fisher vector models. We discuss this choice in Appendix~\ref{app:fisher}.}. Figure~\ref{fig:dominance} shows the dominance probabilities of the ImageNet models as a function of their accuracy differences. Note that these probabilities are less than $0.09$, and $76\%$ of them are less than $0.05$. Also, the dominance probability decreases as the difference in accuracy increases.
Hence, it seems that on ImageNet the more accurate a model is the more it dominates lower accuracy models.

\section{The size of distribution shift}
\label{sec:size_dist_shift}
In this section we discuss the notion of closeness between $\PP$ and $\QQ$ postulated in Definition~\ref{def:dist}. Of course, for any distributions $\PP$ and $\QQ$ there exist parameters $\delta_1$, $\delta_2$, $\nu_1$ and $\nu_2$ such that \eqref{eq:as_refined} holds. However, this trivial point is not sufficient to justify our assumption. Fortunately, similar notions of closeness between distributions have been used in other contexts. 
\citet{kpotufe2018marginal} used a similar metric to analyze the sample complexity of transfer learning. Coincidentally, inequalities of the form \eqref{as:dom} are also used in differential privacy to compare the distributions of outputs of a randomized algorithm applied to two different databases \cite{dwork2006icalp, dwork2006calibrating,dwork2014algorithmic}.  

An important advantage of requirement~\eqref{eq:as_refined} is that it must hold only for a finite number of events for our analysis to go through. As mentioned in Section~\ref{sec:statement}, the fewer the number of events for which \eqref{eq:as_refined} has to hold, the more permissive the notion of closeness between distributions becomes. 

Another advantage of requirement~\eqref{eq:as_refined} is that for the relevant events $A$, we can estimate $\PP(A)$ and $\QQ(A)$ on CIFAR-10 and ImageNet and find $\delta_1$, $\delta_2$, $\nu_1$, and $\nu_2$ such that \eqref{eq:as_refined} holds. Thus, we can estimate the closeness of $\PP$ and $\QQ$ empirically.  

For any data point a model either classifies it correctly or incorrectly. Hence, given three models there are $8$ possible correctness outcomes. Definition~\ref{def:dist} 
considers only $6$ of the corresponding events. To understand why we do not need to worry about $\QQ(A_i^{-} \cap A_j^{-} \cap A_k^{-})$ and $\QQ(A_i^{+} \cap A_j^{+} \cap A_k^{+})$, note that changes in these two probabilities would only move the three points $(\p{i}, \q{i})$, $(\p{j}, \q{j})$, and $(\p{k}, \q{k})$ up or down equally. Therefore, the two probabilities would have no effect on how closely to a line the three points lie.

In Section~\ref{sec:examples} we explained that $\PP$ and $\QQ$ must be close to guarantee approximate collinearity, but what does it mean for $\PP$ and $\QQ$ to be far apart? According to Definition~\ref{def:dist} there must exist an event among the relevant $6\binom{\nummodels}{3}$ events that does not satisfy \eqref{eq:as_refined}. Nonetheless, our analysis can tolerate some violations of \eqref{eq:as_refined}. To understand why,  suppose we are given four models $f_1$, $f_2$, $f_3$, $f_4$, and suppose \eqref{eq:as_refined} is satisfied by the events defined by $(f_1, f_2, f_4)$ and $(f_1, f_3, f_4)$. Then, we can show that $f_2$ and $f_3$ lie close to the line defined by $f_1$ and $f_4$, without needing the events defined by the triplets $(f_1, f_2, f_3)$ and $(f_2,f_3, f_4)$ to satisfy \eqref{eq:as_refined}. Hence, we can guarantee that all models are approximately collinear even when some of the $6\binom{\nummodels}{3}$ events do not satisfy \eqref{eq:as_refined}. 

Distributional closeness does not necessarily have to be defined in terms of linear bounds. Instead, our analysis can be performed with bounds of the form
$
- g_1(\PP(A)) \leq \QQ(A) - \PP(A) \leq g_2(\PP(A))
$
for some functions $g_1$ and $g_2$. Such a notion of closeness is important in the case of CIFAR-10, since for this dataset the linear bounds require large $\delta_2$ and $\nu_2$, although $|\QQ(A) - \PP(A)|$ is small for all relevant events. We discuss this point further in Appendix~\ref{app:cifar}.

\paragraph{ImageNet and ImageNetV2 are close.} We study $\nummodels = 66$ ImageNet models that were collated and evaluated by \citet{recht2019imagenet}. 
We would like to find parameters $\delta_1$, $\delta_2$, $\nu_1$, $\nu_2$ such that the bounds in~\eqref{eq:as_refined} hold for all necessary events defined by triplets of models when the probabilities are evaluated under the ImageNetV2 and ImageNet distributions. 

We evaluate these probabilities empirically, which will have  some estimation error.  Nonetheless, this exercise reveals what are reasonable values of $\delta_1$, $\delta_2$, $\nu_1$, $\nu_2$ so that ImageNet and ImageNetV2 are $(\delta_1, \delta_2, \nu_1, \nu_2)$-close. 

The blue points shown in Figure~\ref{fig:triplets} represent probabilities of the $\num{274560}$  events defined by model triplets. All blue points lie in a wedge defined by two lines, with slopes $0.69$ and $1.38$ and with $y$-axis intercepts $-0.005$ and $0.008$ respectively. This empirical evaluation suggests that ImageNet and ImageNetV2 are $(0.31, 0.38, 0.005, 0.008)$-close.

Interestingly, if we only require $95\%$ of the $\num{274560}$ points to lie inside the wedge, we can choose $\delta_1 = 0$, $\delta_2 = 0.25$, $\nu_1 = 0.005$, and $\nu_2 = 0.005$. Therefore, the vast majority of points lie in a much smaller wedge than the one considered previously. This observation is valuable for Section~\ref{sec:probit}, where we discuss the linear fit in probit domain.

\section{Main arguments}
\label{sec:main}
In Section~\ref{sec:dominance} we present the crux of our analysis under the simplifying assumption that the set of models is \nameset{}. Then, in Section~\ref{sec:general} we discuss what happens when the dominance probabilities are small, but not necessarily zero. 

\subsection{Proof of Proposition~\ref{prop:main}} 
\label{sec:dominance}

Given an ordered set of models let us consider three models $f_1$, $f_2$, and $f_3$ with $\p{1} \leq \p{2} \leq \p{3}$. Also, let $\ell$ be the line between the points $(\p{1}, \q{1})$ and $(\p{3}, \q{3})$. Then, we upper bound the residual from $(\p{2}, \q{2})$ to $\ell$ in the vertical direction (residuals are always measured along the $y$-axis). 

To express this residual we use the notation introduced in Section~\ref{sec:examples}. Then, 
the line $\ell$ is defined by the equation:
\begin{align*}
\ell(\mu) &= \frac{\q{3} - \q{1}}{\p{3} - \p{1}} (\mu - \p{1}) + \q{1} \\
&=\frac{\xq{23} + \xq{3} - \xq{1} - \xq{12}}{p_{23} + p_{3}} (\mu - p_{123})  + \xq{123}   + \xq{12} + \xq{13}  + \xq{1}. 
\end{align*}
Hence, the residual $r := |\ell(\p{2}) - \q{2}|$ is
\begin{align}
\label{eq:dominance_residual}
\nonumber
r &= \left| \frac{\xq{23} + \xq{3} - \xq{1} - \xq{12}}{p_{23} + p_{3}} p_{23} + \xq{13} + \xq{1} - \xq{23} - \xq{2}
\right| \\
&\leq  \left| \frac{\xq{3}p_{23} - p_3 \xq{23}}{p_{23} + p_{3}}
\right|   +   \left|\frac{p_3}{p_3 + p_{23}} \xq{1} - \frac{p_{23}}{p_3 + p_{23}} \xq{12} + \xq{13}  - \xq{2} \right|.
\end{align}
When $\PP$ and $\QQ$ are $(\delta_1, \delta_2, \nu_1, \nu_2)$-close we can use the upper and lower bounds \eqref{eq:as_refined} on $\xq{1}, \xq{13}, \xq{2}, \xq{12}$, $\xq{3}$, and $\xq{23}$, which yields the first part of Proposition~\ref{prop:main}.  
To show that there is a line $\ell^\prime$ that halves the residuals let us denote by $A$, $B$, and $C$ the three points $(\p{1}, \q{1})$, $(\p{2}, \q{2})$, and $(\p{3}, \q{3})$. Now, instead of the line $\ell$ that passes through $A$ and $C$, we consider the line $\ell^\prime$ defined by the middles points of the segments $AB$ and $BC$. Then the residual $r^\prime$ between the three points and $\ell^\prime$ is equal to $r / 2$, gaining a factor of $2$ over \eqref{eq:basic_bound}. This argument completes the proof of Proposition~\ref{prop:main}.

\subsection{Approximately \nameset{} models}
\label{sec:general}

Now we discuss what happens when the set of models $f_1$, $f_2$, \ldots, $f_\nummodels$ is not \nameset{}, i.e., when the dominance probabilities are allowed to be non-zero. Nevertheless, Figure~\ref{fig:dominance} from Section~\ref{sec:similarity} shows that these probabilities can be assumed to be small (e.g. less than $0.05$). 

We proceed as in the Section~\ref{sec:dominance}, using the same notation. Given three models $f_1$, $f_2$, $f_3$ we show that the three points $(\p{1}, \q{1})$, $(\p{2}, \q{2})$, and $(\p{3}, \q{3})$ are approximately collinear. Now, we have $\p{1} = p_{1} + p_{12} + p_{13} + p_{123}$, $\p{2} = p_{2} + p_{12} + p_{23} + p_{123}$, and $\p{3} = p_{3} + p_{13} + p_{23} + p_{123}$. Also, since the set of models satisfies Assumption~\ref{as:dom} we know that $p_1 + p_{12}$, $p_1 + p_{13}$, and $p_2 + p_{12}$ are at most $\zeta$. 

As before, we consider the line $\ell$ defined by $(\p{2}, \q{2})$ and $(\p{3}, \q{3})$:
\begin{align*}
\ell(\mu) = \frac{\q{3} - \q{1}}{\p{3} - \p{1}}(\mu - \p{1}) + \q{1}.
\end{align*}
Therefore, the residual $r$ from $(\p{2}, \q{2})$ to $\ell$ is 
\begin{align}
r 
&= \left| 
\frac{\xq{3} + \xq{23} - \xq{1} - \xq{12}}{p_3 + p_{23} - p_{1} - p_{12}}(p_2 + p_{23} - p_1 - p_{13}) + \xq{1} + \xq{13} - \xq{2} -\xq{23}\right|.
\label{eq:all_terms}
\end{align}

\paragraph{Numerical upper bound.} Given Assumption~\ref{as:dom} and the fact that $\PP$ and $\QQ$ are $(\delta_1, \delta_2, \nu_1, \nu_2)$-close, to get the best possible upper bound on \eqref{eq:all_terms}, one would have to consider many cases, which are determined by the signs of various quantities inside the absolute value. Instead, we compute an upper bound on \eqref{eq:all_terms} numerically.

Given three models $f_1$, $f_2$, $f_3$ that satisfy Assumption~\ref{as:dom} we compute numerically an upper bound on the residual from $f_2$ to the line defined by the other two models. To achieve this we grid over the six probabilities of $p_{1}$, $p_2$, $p_3$, $p_{12}$, $p_{13}$, $p_{23}$. To ensure that Assumption~\ref{as:dom} holds and to ensure that the models have the desired accuracies we impose the following 
constraints: $p_1 + p_{12} \leq \zeta$, $p_{1} + p_{13} \leq \zeta$, $p_{2} + p_{12} \leq \zeta$, and 
\begin{align*}
p_{2} + p_{23} - p_1 - p_{13} &= \p{2} - \p{1},\\
p_{3} + p_{13} - p_2 - p_{12} &= \p{3} - \p{2}.
\end{align*}
These two conditions ensure that we can choose $p_{123}$ so that the models have the desired accuracies. 

Once the $\PP$ probabilities of the events $A_i^{\varepsilon_i} \cap A_j^{{}\varepsilon_j} \cap A_k^{\varepsilon_k}$ are fixed, we grid over the $\QQ$ probabilities, ensuring that
$
\max\{0, -\nu_1 + (1 - \delta_1) p \} \leq \xq{} \leq \nu_2 + (1 + \delta_2)p,
$
for all $p_\cdot \in\{p_1, p_2, p_3, p_{12}, p_{13}, p_{23}\}$ and corresponding $\xq{}$. For each probability we choose five equally-spaced grid points in the relevant interval. 

If we set $\delta_1 = 0.31$, $\delta_2 = 0.38$, $\nu_1 = 0.005$, $\nu_2 = 0.008$, $\zeta = 0.05$ and consider models that are $0.6$, $0.7$, $0.8$ accurate, we find that \eqref{eq:all_terms} is at most $0.084$. Then, by shifting the line $\ell$ towards $(\p{2}, \q{2})$ we can halve the residual. Therefore, in the case we are analyzing, there exists a line such that the residuals of the models $f_1$, $f_2$, and $f_3$ are at most $0.042$. If we choose $\delta_1 = 0.0$, $\delta_2 = 0.25$, $\nu_1 = 0.005$, and $\nu_2 = 0.005$, numerical evaluation shows that \eqref{eq:all_terms} is at most $0.045$, whereby there exists a line such that the residuals of the three models are at most $0.0225$. We also present our numerical bounds in Figures~\ref{fig:probita} and \ref{fig:probitb}, where we also vary $\zeta$. These numerical bounds show that the analytical guarantees derived in Proposition~\ref{prop:main} and Corollary~\ref{cor:main} do not degrade significantly when the set of models is not ordered, but satisfies Assumption~\ref{as:dom} with small $\zeta$.

\subsection{Discussion of probit axis scaling}
\label{sec:probit}

\begin{figure}[t]
\centering
\begin{subfigure}[b]{.45\linewidth}
\includegraphics[width=\linewidth]{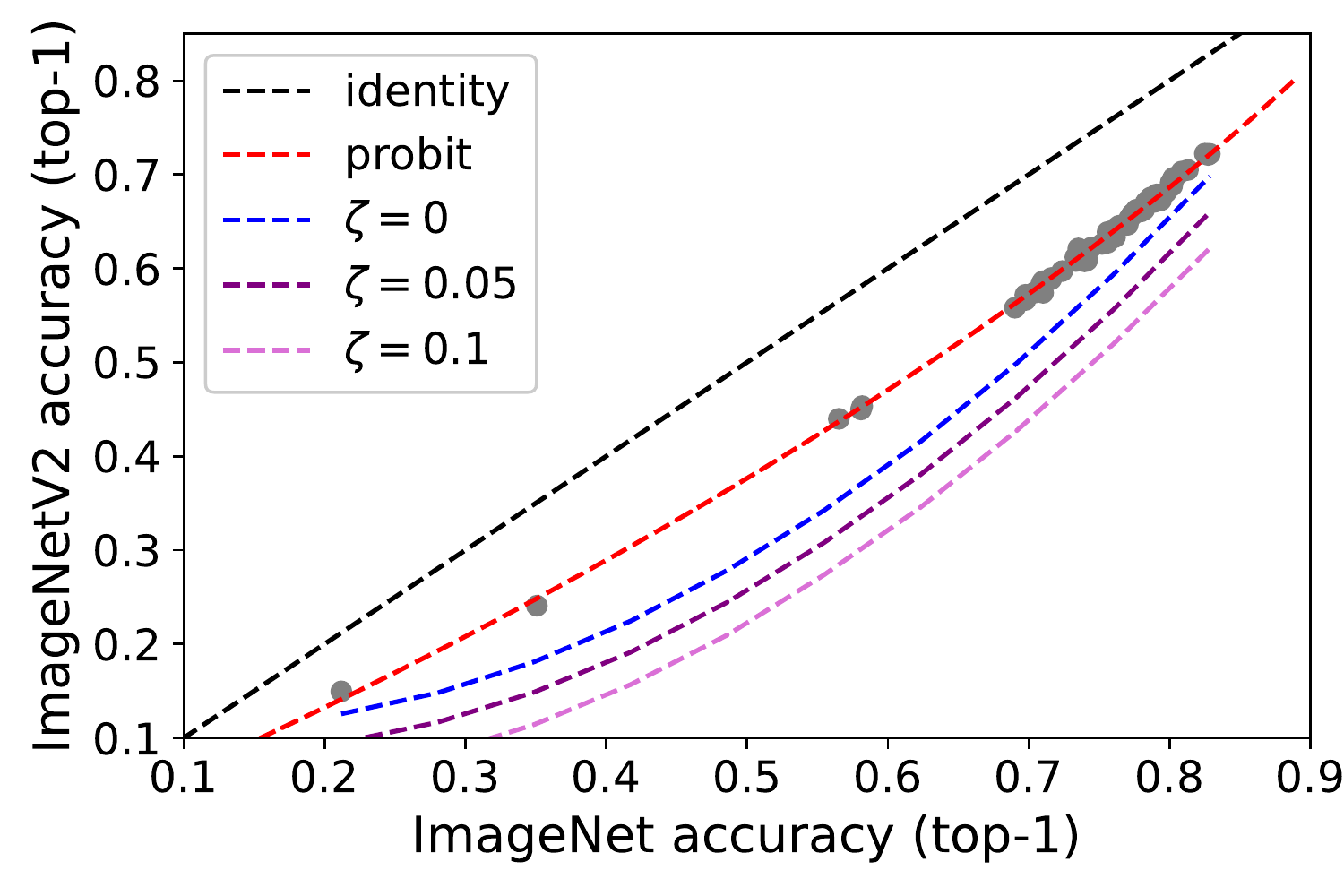}
\caption{Guarantees with $\delta_1 = 0.31$, $\delta_2 = 0.38$ and \\ $\nu_1 = 0.005$, $\nu_2 = 0.008$.}
\label{fig:probita}
\end{subfigure}
\begin{subfigure}[b]{.45\linewidth}
\includegraphics[width=\linewidth]{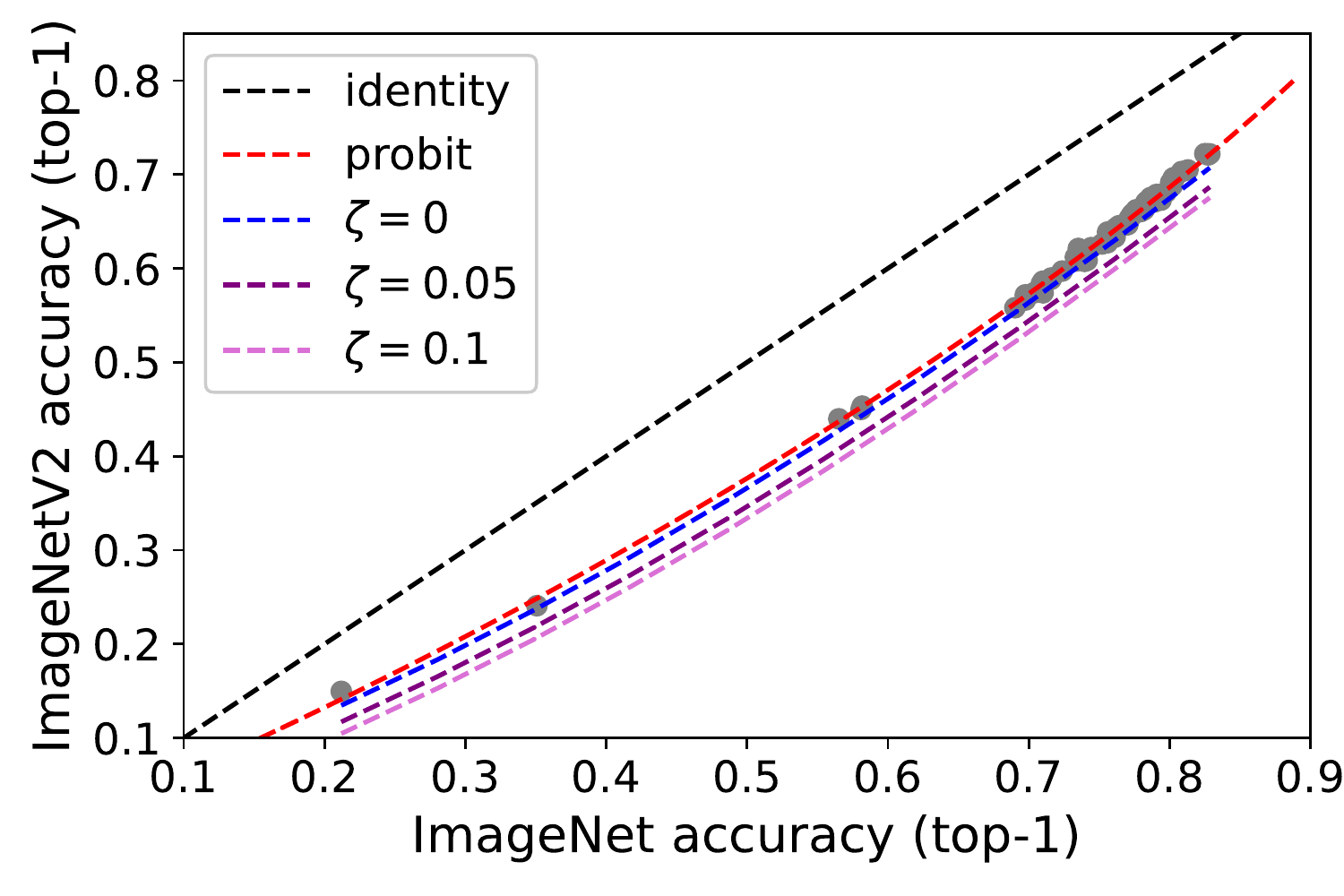}
\caption{Guarantees with $\delta_1 = 0.0$, $\delta_2 = 0.25$, and \\ $\nu_1 = \nu_2 = 0.005$.}
\label{fig:probitb}
\end{subfigure}
\caption{Comparison between probit linear fit of the ImageNet models and our lower bounds on ImageNetV2 accuracies.}
\label{fig:probit}
\end{figure}

\citet{recht2019imagenet} show that  using a probit axes scaling leads to a better linear fit for ImageNet models. Concretely, they fit a line to the points $(\Phi^{-1}(\p{i}), \Phi^{-1}(\q{i}))$, where $\Phi$ is the CDF of the standard normal distribution, and observe a better linear fit than in Figure~\ref{fig:example}. The standard linear fit, shown on the right of Figure~\ref{fig:example}, produces a line with a maximum residual of $0.044$ and a $R^2$ of $0.988$. On the other hand, the linear fit in probit domain, shown as the red dashed curves in Figures~\ref{fig:probita} and \ref{fig:probitb}, yields a maximum residual of $0.01$ and a $R^2$ of $0.998$. 

Given the goodness of this fit, it is natural to wonder whether there is an underlying reason for it. \citet{recht2019imagenet} describe a generative model for the data distributions and models that leads to a linear fit in probit domain. Our assumptions are less restrictive and we use them to explain why it is likely to a have a good fit in probit domain. 

In Figure~\ref{fig:probita}, the red dashed curve is obtained by mapping the linear fit in probit domain back to probability space. To understand what our analysis says about this phenomenon we also show the guarantees provided by \eqref{eq:basic_bound} and \eqref{eq:all_terms} as the blue and purple curves when $\delta_1 = 0.31$, $\delta_2 = 0.38$, $\nu_1 = 0.005$, and $\nu_2 = 0.008$ (values that guarantee that ImageNet and ImageNetV2 are $(\delta_1, \delta_2, \nu_1, \nu_2)$-close). The maximal value of \eqref{eq:all_terms} is computed numerically by gridding, as discussed in Section~\ref{sec:general}. Both the blue and purple curves are lower bounds on test accuracies computed on ImageNetV2. The blue curve assumes the set of models is \nameset{} (i.e. $\zeta  = 0$). The other two curves uses Assumption~\ref{as:dom} with $\zeta = 0.05$ and $\zeta = 0.1$, respectively.  

While the blue and purple curves shown in Figure~\ref{fig:probita} follow loosely the probit (red) curve, they are not tight lower bounds. We can do better. We already discussed that the choice of parameters $\delta_1 = 0.31$, $\delta_2 = 0.38$, $\nu_1 = 0.005$, $\nu_2 = 0.008$ is conservative. If we set $\delta_1 = 0.$, $\delta_2 = 0.25$, and $\nu_1 =\nu_2 = 0.005$, we get tighter bounds that we plot in Figure~\ref{fig:probitb}. Interestingly, the two lower bounds closely follow the probit curve.
Therefore, the goodness of the probit linear fit might be a fortuitous consequence of the models saturating
our bounds. 

\begin{figure}[t]
\centering
\includegraphics[width=0.45\linewidth]{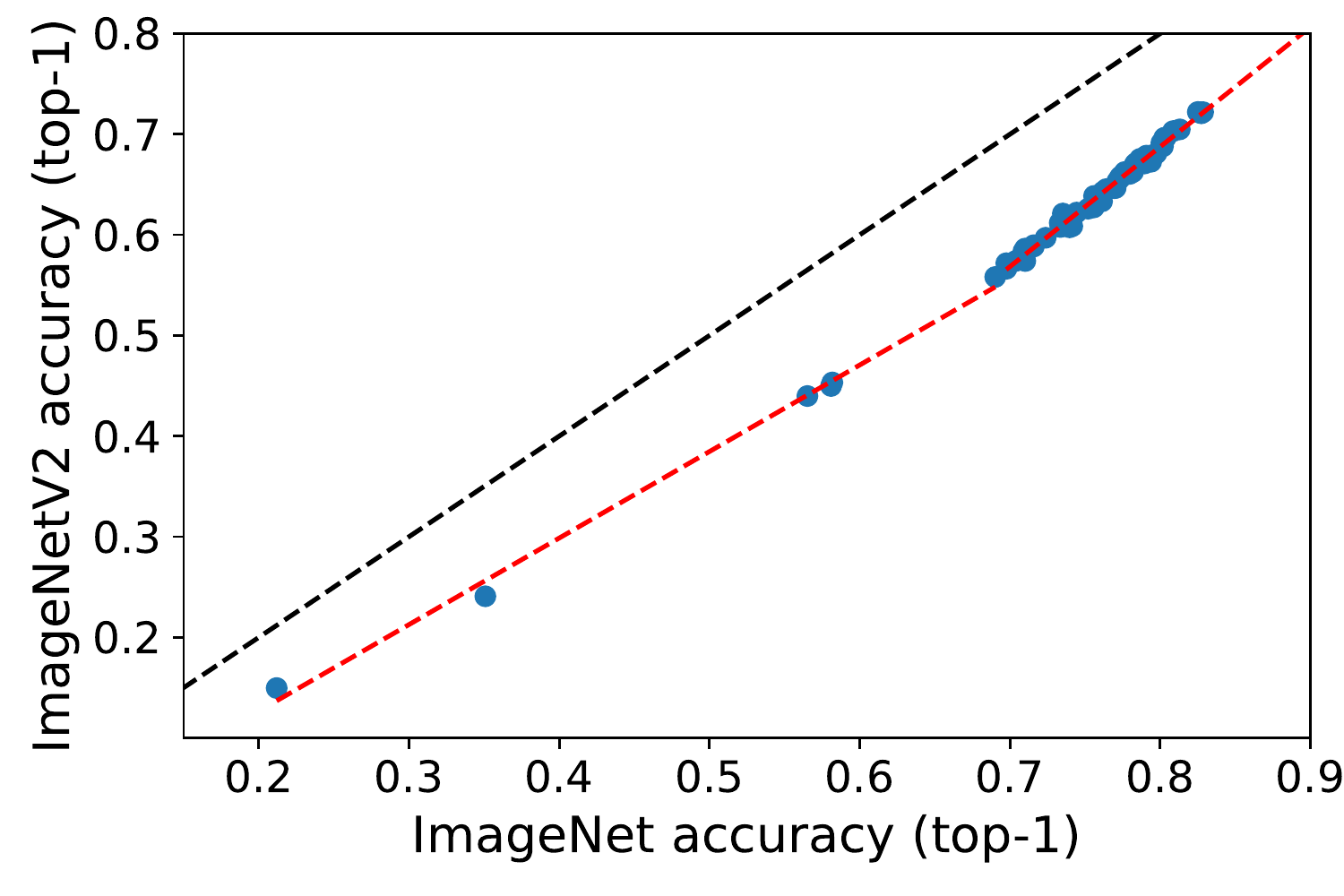}
\caption{Piecewise linear fit of $66$ ImageNet models with two pieces. The switch point between pieces is set to be the $6$-th least accurate model.}
\label{fig:piecewise} 
\end{figure}

Moreover, we said in Section~\ref{sec:statement} that when the upper bound~\eqref{eq:basic_bound} is saturated by $f_1$, $f_\nummodels$, and some third model $f_i$, we expect a piecewise linear function to fit the data much better than a simple linear fit. This conclusion can be reached by applying \eqref{eq:basic_bound} to multiple triplets of models. Indeed, Figure~\ref{fig:piecewise} shows a piecewise linear fit with two pieces. The $R^2$ of this fit is $0.997$, which is almost identical to the $R^2$ of the probit linear fit (the difference between the two $R^2$ scores is $4 \times 10^{-4}$). Observing a good piecewise linear fit is likely to be a more general occurrence than a linear fit in probit domain. In fact, \citet{taori2020measuring} saw that a piecewise linear function fits models fairly well when the models are evaluated on ImageNet-A (a collection of ImageNet test images collected by \citet{hendrycks2019natural}) and the original ImageNet validation set. 

Therefore, while the generative model of \citet{recht2019imagenet} shows that models can be exactly collinear in probit domain, our analysis shows that trends that look linear in probit domain occur more generally. 
Moreover, we saw that a piecewise linear fit with two pieces fits the data equally well. Determining whether a piecewise linear trend occurs more generally than a probit linear trend would be an interesting question for future work to explore.

\section{Conclusion}
\label{sec:discussion}

We said that understanding why models are approximately collinear may lead to insights into improving model robustness. According to our analysis, there are roughly three ways in which a new model can be more robust than previous models. In the first case, the linear fit of the previous models has slope larger than one and the new model has higher accuracy than previous models, but it still lies close to the same line. 
According to terminology introduced by \citet{taori2020measuring}, in this situation the new model has \emph{relative robustness}, but does not have \emph{effective robustness} because it is not more robust than what the linear trend predicts. 

In the second case, even though the distributions $\PP$ and $\QQ$ might be $(\delta_1, \delta_2, \nu_1, \nu_2)$-close with respect to the previous models, the addition of a new model might reveal that the distributions are not close. In this case, some of the relevant events on which the new model classifies correctly can have much higher probabilities under $\QQ$ than under $\PP$ (i.e. $\QQ(A) > \nu_2 + (1 + \delta_2)\PP(A)$), making the new model effectively robust. This situation might occur when the new model is trained on a different dataset than previous models. 

Interestingly, the newly released CLIP model, developed by \citet{radford2learning} to showcase the power of natural language supervision for image classification, is the only model that exhibits improved robustness on ImageNetV2. CLIP was trained on $400$ million images with captions collected from various sources. For context, the ImageNet training set contains one million images \cite{deng2009imagenet}. Therefore it is possible that including CLIP into our set of models would reveal that 
ImageNetV2 and ImageNet are not $(0.31, 0.38, 0.005, 0.008)$-close anymore. We note that there are multiple other models trained on larger datasets, but those models do not exhibit significant robustness with respect to ImageNetV2 \cite{taori2020measuring}.

In the third case, the new model is not similar to previous models. When this happens the new model can be above or below the line described by the previous models even when the two distributions are close. 
Our guess is that CLIP is not similar to previous models which makes effective robustness possible. We leave an analysis of CLIP to future work.

Several open questions remain. We offered sufficient conditions to guarantee that classifications models are approximately collinear, but it is not clear whether there are practical situations in which models are approximately collinear and our conditions are violated. Moreover, it is not clear when we can expect our conditions to hold. In particular, Assumption~\ref{as:dom} is related to model similarity, but we do not understand why models are similar. Are models similar due to the training data, the model classes, or something else? 

\paragraph{Acknowledgments.} We thank Sara Fridovich-Keil, Benjamin Recht, Ludwig Schmidt, and Vaishaal Shankar for valuable feedback that helped us improve the clarity of the manuscript. 

\bibliographystyle{abbrvnat}   
\bibliography{fit}

\begin{thebibliography}{24}
\providecommand{\natexlab}[1]{#1}
\providecommand{\url}[1]{\texttt{#1}}
\expandafter\ifx\csname urlstyle\endcsname\relax
  \providecommand{\doi}[1]{doi: #1}\else
  \providecommand{\doi}{doi: \begingroup \urlstyle{rm}\Url}\fi

\bibitem[Barbu et~al.(2019)Barbu, Mayo, Alverio, Luo, Wang, Gutfreund,
  Tenenbaum, and Katz]{barbu2019objectnet}
A.~Barbu, D.~Mayo, J.~Alverio, W.~Luo, C.~Wang, D.~Gutfreund, J.~Tenenbaum, and
  B.~Katz.
\newblock Objectnet: A large-scale bias-controlled dataset for pushing the
  limits of object recognition models.
\newblock In \emph{Advances in Neural Information Processing Systems}, pages
  9453--9463, 2019.

\bibitem[Ben-Tal et~al.(2013)Ben-Tal, Den~Hertog, De~Waegenaere, Melenberg, and
  Rennen]{ben2013robust}
A.~Ben-Tal, D.~Den~Hertog, A.~De~Waegenaere, B.~Melenberg, and G.~Rennen.
\newblock Robust solutions of optimization problems affected by uncertain
  probabilities.
\newblock \emph{Management Science}, 59\penalty0 (2):\penalty0 341--357, 2013.

\bibitem[Delage and Ye(2010)]{delage2010distributionally}
E.~Delage and Y.~Ye.
\newblock Distributionally robust optimization under moment uncertainty with
  application to data-driven problems.
\newblock \emph{Operations research}, 58\penalty0 (3):\penalty0 595--612, 2010.

\bibitem[Deng et~al.(2009)Deng, Dong, Socher, Li, Li, and
  Fei-Fei]{deng2009imagenet}
J.~Deng, W.~Dong, R.~Socher, L.-J. Li, K.~Li, and L.~Fei-Fei.
\newblock Image{N}et: A large-scale hierarchical image database.
\newblock In \emph{2009 IEEE conference on computer vision and pattern
  recognition}, pages 248--255. IEEE, 2009.

\bibitem[Deza and Deza(2018)]{deza}
M.~M. Deza and E.~Deza.
\newblock \emph{Encyclopedia of Distances}.
\newblock Springer, second edition, 2018.

\bibitem[Duchi et~al.(2020)Duchi, Hashimoto, and
  Namkoong]{duchi2019distributionally}
J.~C. Duchi, T.~Hashimoto, and H.~Namkoong.
\newblock Distributionally robust losses against mixture covariate shifts.
\newblock \emph{arXiv:2007.13982}, 2020.

\bibitem[Dwork(2006)]{dwork2006icalp}
C.~Dwork.
\newblock Differential privacy.
\newblock In \emph{Proceedings of the International Colloquium on Automata,
  Languages and Programming (ICALP)}, pages 1--12, 2006.

\bibitem[Dwork et~al.(2006)Dwork, McSherry, Nissim, and
  Smith]{dwork2006calibrating}
C.~Dwork, F.~McSherry, K.~Nissim, and A.~Smith.
\newblock Calibrating noise to sensitivity in private data analysis.
\newblock In \emph{Theory of cryptography conference}, pages 265--284.
  Springer, 2006.

\bibitem[Dwork et~al.(2014)Dwork, Roth, et~al.]{dwork2014algorithmic}
C.~Dwork, A.~Roth, et~al.
\newblock The algorithmic foundations of differential privacy.
\newblock \emph{Foundations and Trends in Theoretical Computer Science},
  9\penalty0 (3-4):\penalty0 211--407, 2014.

\bibitem[Esfahani and Kuhn(2018)]{esfahani2018data}
P.~M. Esfahani and D.~Kuhn.
\newblock Data-driven distributionally robust optimization using the
  {W}asserstein metric: {P}erformance guarantees and tractable reformulations.
\newblock \emph{Mathematical Programming}, 171\penalty0 (1-2):\penalty0
  115--166, 2018.

\bibitem[Hendrycks et~al.(2019)Hendrycks, Zhao, Basart, Steinhardt, and
  Song]{hendrycks2019natural}
D.~Hendrycks, K.~Zhao, S.~Basart, J.~Steinhardt, and D.~Song.
\newblock Natural adversarial examples.
\newblock \emph{arXiv preprint arXiv:1907.07174}, 2019.

\bibitem[Kpotufe and Martinet(2018)]{kpotufe2018marginal}
S.~Kpotufe and G.~Martinet.
\newblock Marginal singularity, and the benefits of labels in covariate-shift.
\newblock In \emph{Proceedings of the 31st Conference On Learning Theory},
  volume~75 of \emph{Proceedings of Machine Learning Research}, pages
  1882--1886. PMLR, 2018.

\bibitem[Krizhevsky(2009)]{krizhevsky2009learning}
A.~Krizhevsky.
\newblock Learning multiple layers of features from tiny images.
\newblock 2009.

\bibitem[Mania et~al.(2019)Mania, Miller, Schmidt, Hardt, and
  Recht]{mania2019model}
H.~Mania, J.~Miller, L.~Schmidt, M.~Hardt, and B.~Recht.
\newblock Model similarity mitigates test set overuse.
\newblock In \emph{Advances in Neural Information Processing Systems}, pages
  9993--10002, 2019.

\bibitem[Miller et~al.(2020)Miller, Krauth, Recht, and
  Schmidt]{miller2020effect}
J.~Miller, K.~Krauth, B.~Recht, and L.~Schmidt.
\newblock The effect of natural distribution shift on question answering
  models.
\newblock In \emph{Proceedings of the 37th International Conference on Machine
  Learning}, volume 119 of \emph{Proceedings of Machine Learning Research},
  pages 6905--6916. PMLR, 2020.

\bibitem[Radford et~al.(2021)Radford, Kim, Hallacy, Ramesh, Goh, Agarwal,
  Sastry, Askell, Mishkin, Clark, et~al.]{radford2learning}
A.~Radford, J.~W. Kim, C.~Hallacy, A.~Ramesh, G.~Goh, S.~Agarwal, G.~Sastry,
  A.~Askell, P.~Mishkin, J.~Clark, et~al.
\newblock Learning transferable visual models from natural language
  supervision.
\newblock \emph{OpenAI}, 2021.

\bibitem[Recht et~al.(2019)Recht, Roelofs, Schmidt, and
  Shankar]{recht2019imagenet}
B.~Recht, R.~Roelofs, L.~Schmidt, and V.~Shankar.
\newblock Do {I}mage{N}et classifiers generalize to {I}mage{N}et?
\newblock In \emph{Proceedings of the 36th International Conference on Machine
  Learning}, volume~97 of \emph{Proceedings of Machine Learning Research},
  pages 5389--5400. PMLR, 2019.

\bibitem[Roelofs et~al.(2019)Roelofs, Shankar, Recht, Fridovich-Keil, Hardt,
  Miller, and Schmidt]{roelofs2019meta}
R.~Roelofs, V.~Shankar, B.~Recht, S.~Fridovich-Keil, M.~Hardt, J.~Miller, and
  L.~Schmidt.
\newblock A meta-analysis of overfitting in machine learning.
\newblock In \emph{Advances in Neural Information Processing Systems}, pages
  9179--9189, 2019.

\bibitem[Sagawa et~al.(2019)Sagawa, Koh, Hashimoto, and
  Liang]{sagawa2019distributionally}
S.~Sagawa, P.~W. Koh, T.~B. Hashimoto, and P.~Liang.
\newblock Distributionally robust neural networks for group shifts: {O}n the
  importance of regularization for worst-case generalization.
\newblock \emph{arXiv preprint arXiv:1911.08731}, 2019.

\bibitem[Shafieezadeh~Abadeh et~al.(2015)Shafieezadeh~Abadeh,
  Mohajerin~Esfahani, and Kuhn]{shafieezadeh2015distributionally}
S.~Shafieezadeh~Abadeh, P.~M. Mohajerin~Esfahani, and D.~Kuhn.
\newblock Distributionally robust logistic regression.
\newblock \emph{Advances in Neural Information Processing Systems},
  28:\penalty0 1576--1584, 2015.

\bibitem[Shankar et~al.(2020)Shankar, Roelofs, Mania, Fang, Recht, and
  Schmidt]{shankar2020evaluating}
V.~Shankar, R.~Roelofs, H.~Mania, A.~Fang, B.~Recht, and L.~Schmidt.
\newblock Evaluating machine accuracy on {I}mage{N}et.
\newblock In \emph{Proceedings of the 37th International Conference on Machine
  Learning}, volume 119 of \emph{Proceedings of Machine Learning Research},
  pages 8634--8644. PMLR, 2020.

\bibitem[Sinha et~al.(2018)Sinha, Namkoong, Volpi, and
  Duchi]{sinha2017certifying}
A.~Sinha, H.~Namkoong, R.~Volpi, and J.~Duchi.
\newblock Certifying some distributional robustness with principled adversarial
  training.
\newblock \emph{International Conference on Learning Representations}, 2018.

\bibitem[Taori et~al.(2020)Taori, Dave, Shankar, Carlini, Recht, and
  Schmidt]{taori2020measuring}
R.~Taori, A.~Dave, V.~Shankar, N.~Carlini, B.~Recht, and L.~Schmidt.
\newblock Measuring robustness to natural distribution shifts in image
  classification.
\newblock In \emph{Advances in Neural Information Processing Systems}, 2020.

\bibitem[Yadav and Bottou(2019)]{yadav2019cold}
C.~Yadav and L.~Bottou.
\newblock Cold case: The lost {MNIST} digits.
\newblock In \emph{Advances in Neural Information Processing Systems}, pages
  13443--13452, 2019.

\end{thebibliography}

\newpage
\appendix 

\section{CIFAR-10.1 vs CIFAR-10}
\label{app:cifar}

In this section we compare CIFAR-10.1 and CIFAR-10 in terms of $(\delta_1, \delta_2, \nu_1, \nu_2)$-closeness based on 30 models collated by \citet{recht2019imagenet}.

\begin{figure}[h]
\begin{subfigure}[b]{.45\linewidth}
\includegraphics[width=\linewidth]{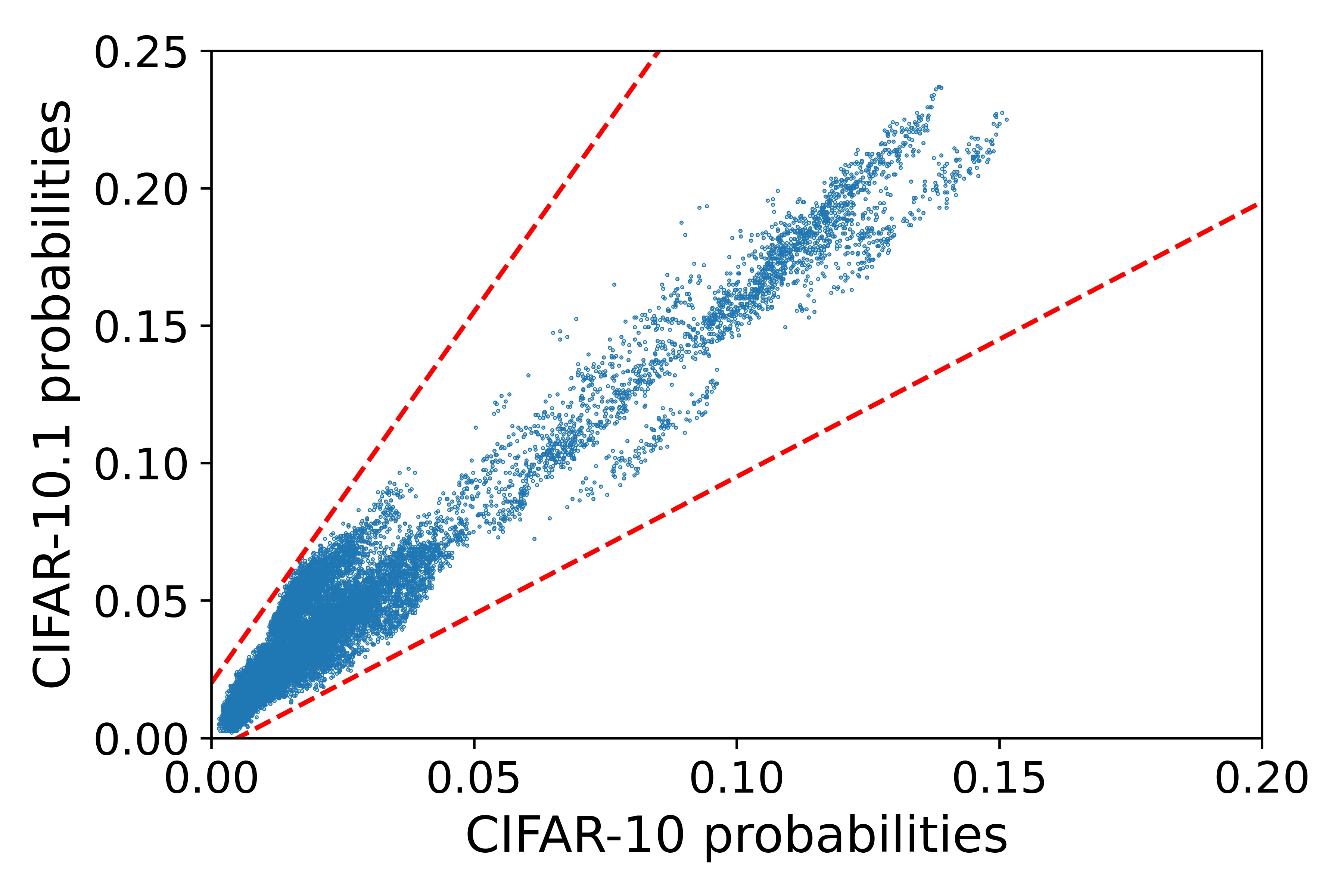}
\caption{Events defined by triplets}
\label{fig:cifar_triplets}
\end{subfigure}
\begin{subfigure}[b]{.45\linewidth}
\includegraphics[width=\linewidth]{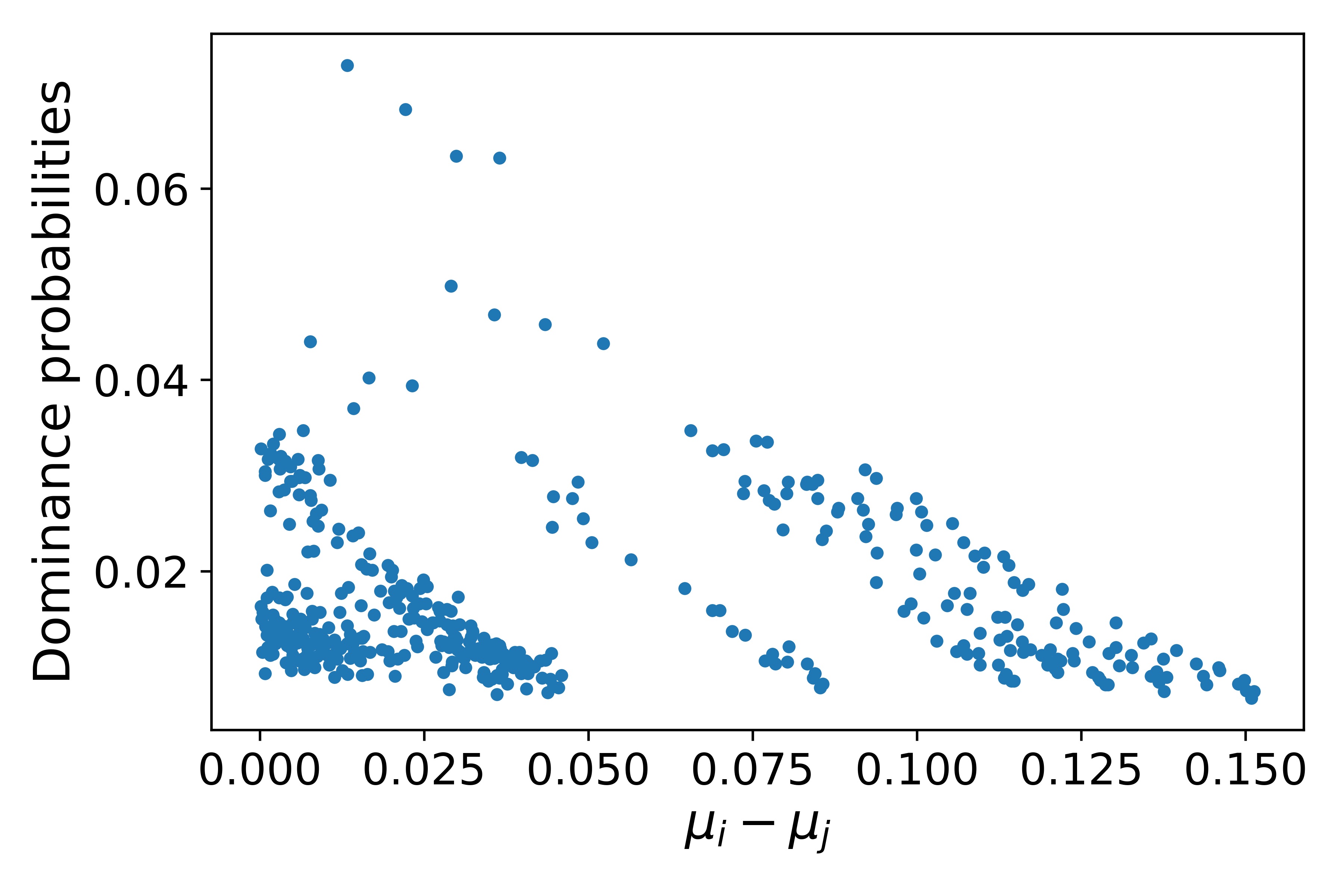}
\caption{Dominance probabilities}
\label{fig:cifar_dominance}
\end{subfigure}
\caption{Figure~\ref{fig:triplets} shows $\num{24360}$ blue points representing the probabilities of the events $A_i^{\varepsilon_i} \cap A_j^{\varepsilon_j} \cap A_k^{\varepsilon_k}$. 
The red dotted lines are upper and lower bounds on the probabilities under CIFAR-10.1, as in \eqref{eq:as_refined}, with parameters: $\delta_1 = 0.0$, $\delta_2 = 1.7$, $\nu_1 = 0.005$, and $\nu_2 = 0.02$. Figure~\ref{fig:dominance} shows the probabilities of the events $A_i^+ \cap A_j^-$ for all pairs of models with $\p{i} < \p{j}$ (i.e., the events on which a worse model is correct, but a better model is incorrect).}
\label{fig:as_evidence_cifar} 
\end{figure}

The blue points shown in Figure~\ref{fig:cifar_triplets} represent probabilities of the $\num{274560}$  events defined by the triplets $\{f_i, f_j, f_k\}$, as considered in Definition~\ref{def:dist}. This empirical evaluation suggests that CIFAR-10 and CIFAR-10.1 satisfy our assumption with $\delta_1 = 0.0$, $\delta_2 = 1.7$, $\nu_1 = 0.005$, and $\nu_2 = 0.02$.

It is important to note that $\delta_2 = 1.7$ is a large value and if we were to plug it into \eqref{eq:basic_bound} we would obtain a poor guarantee. However, we remark that the large slope is needed only for events whose probabilities according to CIFAR-10 are at most $0.03$. 
For events with larger CIFAR-10 probabilities we could use a smaller $\delta_2$. For example, we could upper bound $\QQ(A)$ using $\delta_2 = 0.8$ and $\nu_2 = 0.01$ for all events
considered in Definition~\ref{def:dist} that have $\PP(A) \geq 0.03$. This choice would ensure that \eqref{eq:as_refined} are satisfied for over $95\%$ of events. As explained in Section~\ref{sec:size_dist_shift}, our analysis can tolerate some violations of \eqref{eq:as_refined}. 

For convenience we reiterate our explanation here. Suppose we are given four models $f_1$, $f_2$, $f_3$, $f_4$, and suppose \eqref{eq:as_refined} is satisfied by the events defined by $(f_1, f_2, f_4)$ and $(f_1, f_3, f_4)$. Then, we can show that $f_2$ and $f_3$ lie close to the line defined by $f_1$ and $f_4$, without needing the events defined by the triplets $(f_1, f_2, f_3)$ and $(f_2,f_3, f_4)$ to satisfy \eqref{eq:as_refined}. Hence, we can guarantee that all models are approximately collinear even when some of the $6\binom{\nummodels}{3}$ events do not satisfy \eqref{eq:as_refined}. 

Moreover, we note that distribution closeness does not necessarily have to be defined in terms of linear upper and lower bounds, as in \eqref{eq:as_refined}. Instead, our analysis can be performed with bounds of the form
$
- g_1(\PP(A)) \leq \QQ(A) - \PP(A) \leq g_2(\PP(A))
$
for some nonnegative functions $g_1$ and $g_2$. In other words, our analysis can be carried out whenever $|\PP(A) - \QQ(A)|$ is small for the events considered in Definition~\ref{def:dist}. 

Figure~\ref{fig:cifar_dominance} shows the dominance probabilities of all the pairs of CIFAR-10 models we considers. We note that all these probabilities are smaller than $0.08$, with most of them being at most $0.04$. Therefore, Assumption~\ref{as:dom} is also satisfied with a small $\zeta$. 

\section{Proof of Corollary~\ref{cor:main}}
\label{app:cor}

Let $A = (\p{1}, \q{1})$ and $C = (\p{\nummodels}, \q{\nummodels})$, and let $B = (\p{i}, \q{i})$ be the point that has the largest residual from the line $AC$. Without loss of generality we assume that $B$ is below or on the line $AC$. Then, there are two cases: all other models fall on or below the line $AC$, or there exists at least a point above the line $AC$.

The first case is immediately resolved by applying Proposition~\ref{prop:main} to upper bound the residual $r_B$ of $B$ from the line $AC$. Then, we can choose $\ell$ to be the line that passes through the middle of the segments $AB$ and $BC$. For this choice of $\ell$ the residual of any other point in our collection is upper bounded by $r_B / 2$ because $B$ has the largest residual and all other points lie on or below $AC$. The conclusion follows because the harmonic mean of two numbers with a constant sum is maximized when the two numbers are equal.

For the second case, let $D = (\p{j}, \q{j})$ a point above the line $AC$ with the largest residual from $AC$. We can assume that $\p{1} < \p{j} < \p{i}$. 
Let $r_B$ and $r_D$ be $B$'s and $D$'s residuals from the line $AC$. By our extremal choices of $B$ and $D$ we know that if we consider the lines parallel to $AC$ that pass through $B$ and $D$ respectively, we are guaranteed that all other points lie between them. Therefore, we can choose a line parallel to $AC$ that has residual at most $(r_D + r_B) / 2$ from all points. 

Hence, we are left to upper bound $r_B$ and $r_D$. From Proposition~\ref{prop:main} we know that
\begin{align}
\label{eq:bound_on_B}
r_B \leq \frac{2(\p{\nummodels} - \p{i})(\p{1} - \p{i})}{\p{\nummodels} - \p{1}} \frac{\delta_1 + \delta_2}{2} + 3\max\{\nu_1, \nu_2\}.
%\frac{\q{1} - \q{\nummodels}}{\p{1} - \p{\nummodels}}(\p{i} - \p{\nummodels}) - (\q{i} - \q{\nummodels}). 
\end{align}
Since $D$ is above the line $AC$, if we define by $r_D^\prime$ the residual from $D$ to $BC$, we have $r_D \leq r_D^\prime$. Then, from Proposition~\ref{prop:main} we know that

\begin{align}
\label{eq:bound_on_D}
r_D &\leq r_D^\prime \leq \frac{2(\p{i} - \p{j})(\p{j} - \p{1})}{\p{i} - \p{1}} \frac{\delta_1 + \delta_2}{2}  + 3\max\{\nu_1, \nu_2\}.
\end{align}

Let $c := \p{j} - \p{1}$, $b:= \p{i} - \p{j}$, and $a:= \p{\nummodels} - \p{i}$. Then, putting together \eqref{eq:bound_on_B} and \eqref{eq:bound_on_D} we find 
\begin{align*}
r_B + r_D &\leq \frac{2bc}{b + c} \frac{\delta_1 + \delta_2}{2} + \frac{2a(b + c)}{a + b + c} \frac{\delta_1 + \delta_2}{2}  + 6\max\{\nu_1, \nu_2\}.
\end{align*}
Now, we would like to understand how large can the right hand side be as a function of just $\delta_1$, $\delta_2$, $\nu_1$, $\nu_2$, and $\p{\nummodels} - \p{1}$. In order to do this we find the maximum of the right hand side with respect to $a$, $b$, and $c$ under the constraints $a, b, c \geq 0$ and $a + b + c = \p{\nummodels} - \p{1}$. Using simple first order conditions, one can find that 
\begin{align*}
r_B + r_D &\leq \frac{25}{32} (\p{1} - \p{\nummodels}) \delta + 6\max\{\nu_1, \nu_2\}.
\end{align*}
The result follows by taking $\ell$ to be the line parallel to $AC$ with equal residuals to $B$ and $D$. 
We note that with a more involved proof it is possible to improve the constant $25/64$.

\section{The missing Fisher vector model}
\label{app:fisher}

\citet{recht2019imagenet} collected and evaluated $67$ models in their testbed. However, in the main text we chose to analyze only $66$ of these models. In this section we explain our choice. 

\citet{recht2019imagenet} trained and implemented three Fisher vector models, with $16$, $64$, and $256$ Gaussian Mixture Model centers. In our main analysis we removed the model that uses $64$ GMM centers. We made this choice because leaving in this model would mislead one to think ImageNet and ImageNetV2 are far apart, 
as measured by Definition~\ref{def:dist}. 

\begin{figure}[t]
\centering
\begin{subfigure}[b]{.45\linewidth}
\includegraphics[width=\linewidth]{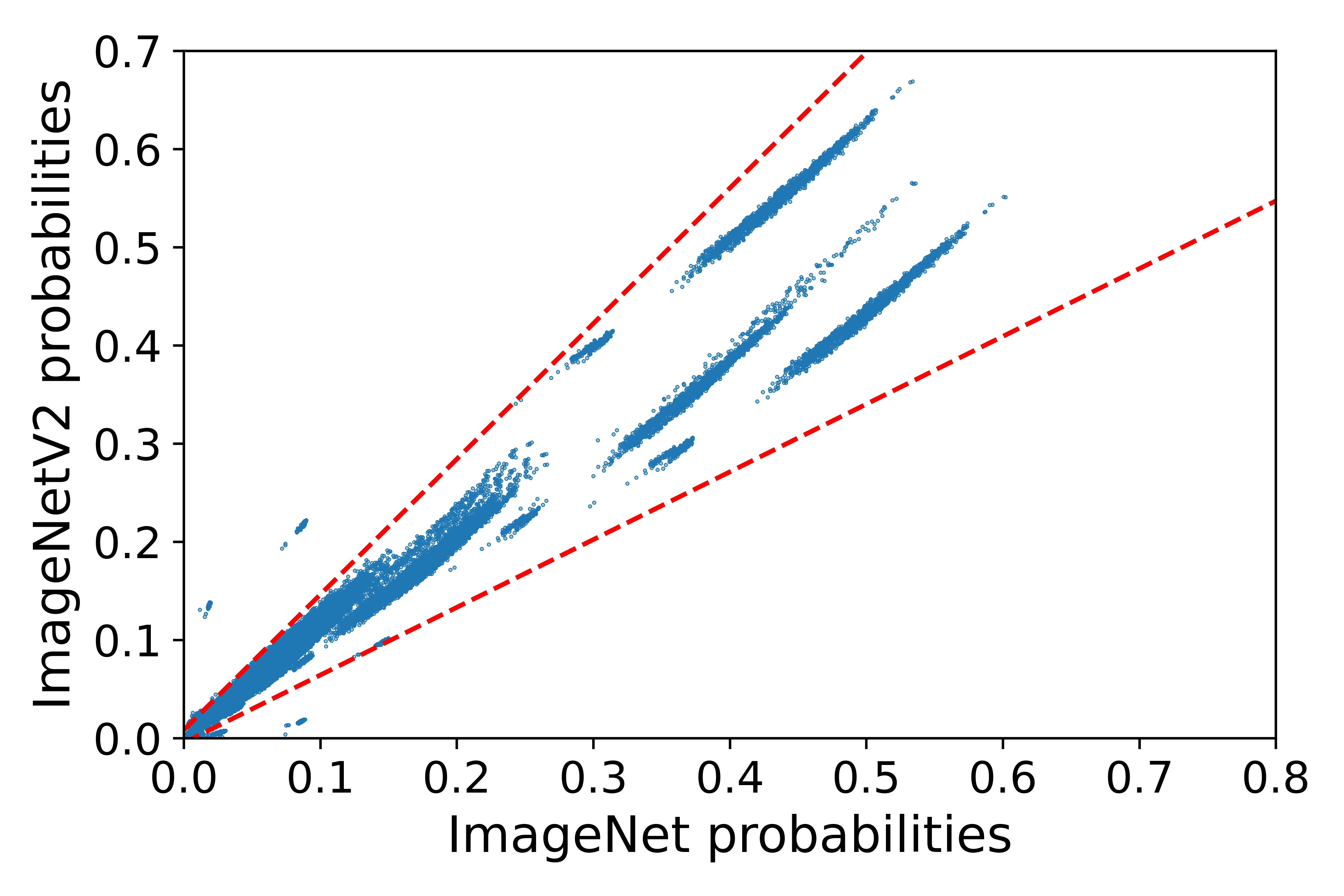}
\caption{Events defined by triplets}
\label{fig:triplets_fisher}
\end{subfigure}
\begin{subfigure}[b]{.45\linewidth}
\includegraphics[width=\linewidth]{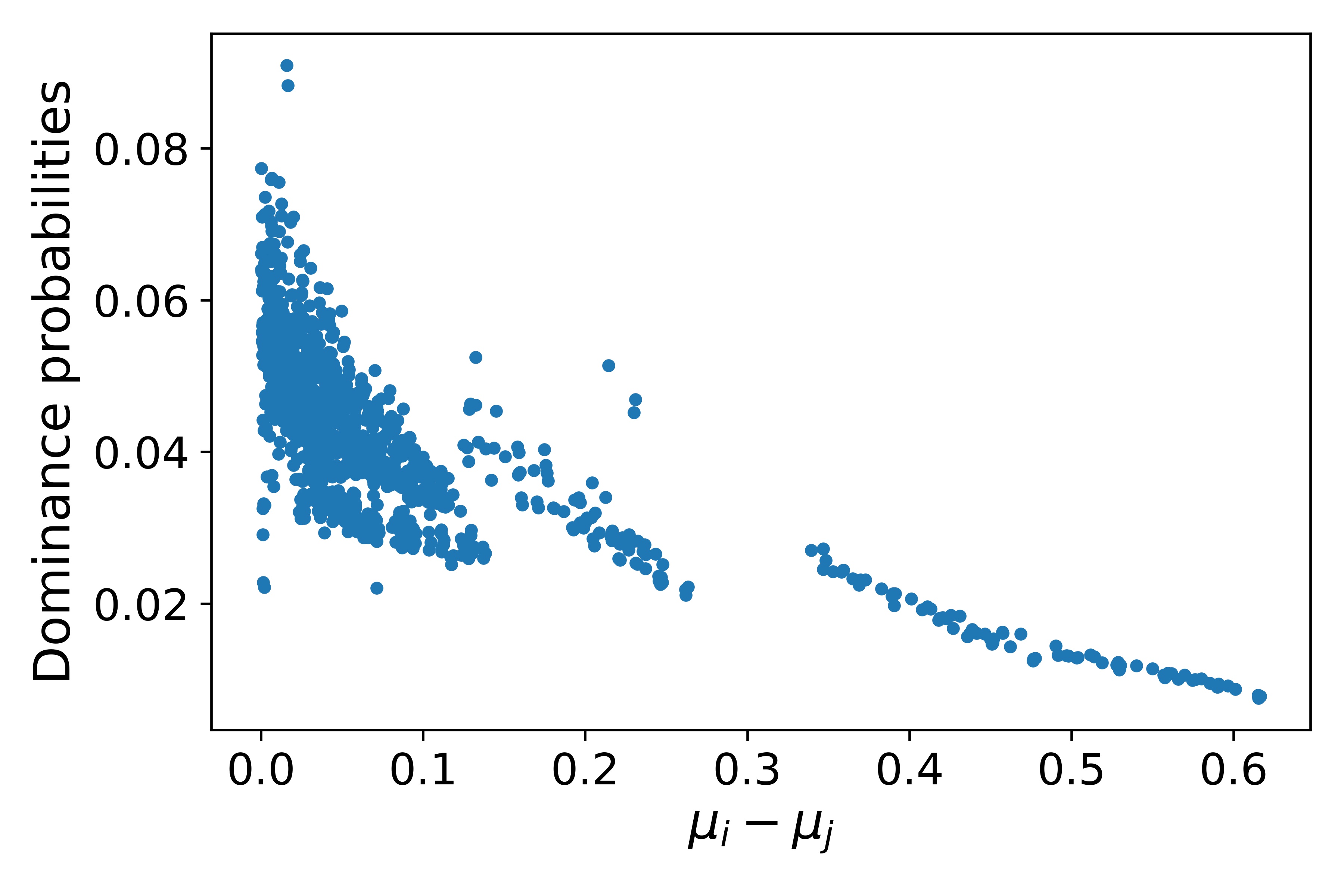}
\caption{Dominance probabilities}
\label{fig:dominance_fisher}
\end{subfigure}
\caption{Figure~\ref{fig:triplets} shows $\num{287430}$ blue points representing the probabilities of the events $A_i^{\varepsilon_i} \cap A_j^{\varepsilon_j} \cap A_k^{\varepsilon_k}$. 
The red dotted lines are upper and lower bounds, as in \eqref{eq:as_refined}, with parameters: $\delta_1 = 0.31$, $\delta_2 = 0.38$, $\nu_1 = 0.005$, and $\nu_2 = 0.008$. Figure~\ref{fig:dominance} shows the dominance probabilities for all pairs of $67$ models.}
\label{fig:as_evidence_fisher} 
\end{figure}

Figure~\ref{fig:triplets_fisher} is a reproduction of Figure~\ref{fig:triplets} with the missing Fisher vector included. We can see that now the wedge defined by $\delta_1 = 0.31$, $\delta_2 = 0.38$, $\nu_1 = 0.005$, $\nu_2 = 0.008$ does not include all the $\num{287430}$ events defined by the $67$ models. 

Since there are $\binom{66}{2} = 2145$ pairs of models that do not include the missing Fisher vector model and since each model triplet defines $6$ relevant events, the missing Fisher vector model introduces $\num{12870}$ new points in Figure~\ref{fig:triplets} compared to Figure~\ref{fig:triplets}. However, just $666$ of these  events lie outside the wedge. Moreover, only $195$ events are included in the three clusters that are far apart from the identity line, with the remaining points falling close to the origin. 

As explained in Section~\ref{sec:size_dist_shift} this kind of violation of \ref{eq:as_refined} does not preclude our analysis. Also, the inclusion of the missing Fisher vector model does not violate Assumption~\ref{as:dom}, as can be seen from Figure~\ref{fig:dominance_fisher}. Hence, for the sake of clarity we chose to omit the problematic Fisher vector model. Further investigation is needed to determine why this particular model defines events whose probabilities shift so much between ImageNet and ImageNetV2. 

\end{document}